\documentclass[a4paper,11pt,dvips,titlepage]{article}
\usepackage{graphicx}
\usepackage{latexsym,amsmath,amssymb,amsthm,color,xypic}
\usepackage{algorithm,algorithmic,subfig}
\usepackage[all]{xy}
\UseCrayolaColors
\CompileMatrices %
\floatstyle{ruled}
\restylefloat{algorithm}
\restylefloat{table}
\restylefloat{figure}

\newtheorem{theorem}{Theorem}

\newtheorem{lemma}[theorem]{Lemma}
\newtheorem{proposition}[theorem]{Proposition}

\theoremstyle{definition}
\newtheorem{definition}[theorem]{Definition}
\newtheorem{example}{Example}[subsection]
\theoremstyle{remark}

\newenvironment{procedure}[1]{\item[#1]}{}

\newcommand{\qedex}{\renewcommand{\qedsymbol}{$\Diamond$}\qed}

\newcommand{\pim}{\ensuremath{\pi_\textsc{\tiny{m}}}}
\newcommand{\pit}{\ensuremath{\pi_\textsc{\tiny{t}}}}
\newcommand{\pik}{\ensuremath{\pi_\textsc{\tiny{k}}}}

\newcommand{\posints}{\mathbb{Z}_+}
\newcommand{\nats}{{\ensuremath{\mathbb{N}}}}

\newcommand{\esp}{\ensuremath{\pi}}         %
\newcommand{\joint}{\ensuremath{P}}         %
\newcommand{\xspace}{{\ensuremath{\cal X}}} %
\newcommand{\rv}[1]{\mbox{\boldmath{$#1$}}}

\newcommand{\prd}{\textnormal{\tiny p}}
\newcommand{\sil}{\textnormal{\tiny s}}
\newcommand{\unf}{\textnormal{\tiny u}}

\newcommand{\name}[2][+<0pt,+10pt>]{\save[]#1*\txt{\tiny$\tuple{\text{#2}}$}\restore}

\newcommand{\wstates}{Q}
\newcommand{\wstate}{q}

\newcommand{\conc}{\smallfrown}

\newcommand{\lemref}[1]{Lemma~\ref{#1}}
\newcommand{\algref}[1]{Algorithm~\ref{#1}}
\newcommand{\thmref}[1]{Theorem~\ref{#1}}
\newcommand{\figref}[1]{Figure~\ref{#1}}
\newcommand{\defref}[1]{Definition~\ref{#1}}
\newcommand{\secref}[1]{Section~\ref{#1}}
\newcommand{\exref}[1]{Example~\ref{#1}}
\newcommand{\dif}{\textnormal{d}}
\newcommand{\wnlsymbol}{\Lambda}
\DeclareMathOperator{\wtf}{P} %
\DeclareMathOperator{\wnl}{\wnlsymbol} %

\newcommand{\winit}{\operatorname{P}_{\!\!\circ}}
\DeclareMathOperator{\simplex}{\triangle}
\DeclareMathOperator{\trace}{\mathit \wnlsymbol}
\DeclareMathOperator{\Alg}{Alg}

\DeclareMathOperator{\Exp}{E}
\DeclareMathOperator{\dom}{dom}

\newcommand{\markdef}[1]{\underline{\textsl{#1}}}

\newcommand{\set}[1]{\left\{#1\right\}}
\newcommand{\del}[1]{\left(#1\right)}
\newcommand{\tuple}[1]{\left\langle#1\right\rangle}
\newcommand{\sbr}[1]{\left[#1\right]}
\newcommand{\card}[1]{\left\lvert#1\right\rvert}

\newcommand{\sw}{\textnormal{sw}}
\newcommand{\A}{\mathbb A}
\newcommand{\argmax}{\mathop{\textnormal{argmax}}}

\newcommand*{\partialfn}{%
  \mathrel{\vcenter{\offinterlineskip
  \hbox{$\cdot$}\vskip-.50ex\hbox{$\cdot$}\vskip-.50ex\hbox{$\cdot$}}}}

\newcommand{\wfrac}[2]{\left. #1 \right/ #2}

\newcommand{\intoo}[2]{\left(#1,#2\right)}
\newcommand{\intoc}[2]{\left(#1,#2\right]}
\newcommand{\intco}[2]{\left[#1,#2\right)}
\newcommand{\intcc}[2]{\left[#1,#2\right]}

\newcommand{\patchbl}[1]{\vbox{#1\hbox{}}}

\newcommand{\expname}[1]{\textsc{#1}}
\newcommand{\ex}[1]{*+[o][F]{\expname{\small#1}}}
\newcommand{\si}{*+<4pt>[o][F**:Black]{}} %
\newcommand{\bn}{*+<4pt>[o][F]{}} %

\newcommand{\family}[2]{\tuple{#1}_{#2}}

\author{Wouter Koolen \and Steven de Rooij}
\title{\scshape Combining Expert Advice Efficiently}
\date{\scriptsize Centrum voor Wiskunde en Informatica (CWI)\\
  Kruislaan 413, P.O. Box 94079\\
  1090 GB Amsterdam, The Netherlands\\
  \ttfamily \{W.M.Koolen-Wijkstra,S.de.Rooij\}@cwi.nl
}

\begin{document}
\maketitle

\tableofcontents

\begin{abstract}
  We show how models for prediction with expert advice can be defined
  concisely and clearly using hidden Markov models (HMMs); standard
  HMM algorithms can then be used to efficiently calculate, among
  other things, how the expert predictions should be weighted
  according to the model. We cast many existing models as HMMs and
  recover the best known running times in each case. We also describe
  two new models: the switch distribution, which was recently
  developed to improve Bayesian/Minimum Description Length model
  selection, and a new generalisation of the fixed share algorithm
  based on run-length coding. We give loss bounds for all models and
  shed new light on their relationships.
\end{abstract}

\section{Introduction}\label{sec:intro}
We cannot predict exactly how complicated processes such as the
weather, the stock market, social interactions and so on, will develop
into the future. Nevertheless, people do make weather forecasts and
buy shares all the time. Such predictions can be based on formal
models, or on human expertise or intuition. An investment company may
even want to choose between portfolios on the basis of a combination
of these kinds of predictors. In such scenarios, predictors typically
cannot be considered ``true''. Thus, we may well end up in a position
where we have a whole collection of prediction strategies, or
\emph{experts}, each of whom has \emph{some} insight into \emph{some}
aspects of the process of interest. We address the question how a
given set of experts can be combined into a single predictive strategy
that is as good as, or if possible even better than, the best
individual expert.

The setup is as follows. Let $\Xi$ be a finite set of experts. Each
expert $\xi\in\Xi$ issues a distribution $P_\xi(\rv{x}_{n+1}|x^n)$ on
the next outcome $\rv{x}_{n+1}$
given the previous observations $x^n:=x_1,\ldots,x_n$. Here, each
outcome $x_i$ is an element of some countable space $\xspace$, and random variables are written in bold face. The
probability that an expert assigns to a sequence of outcomes is given
by the chain rule: $P_\xi(x^n)=P_\xi(x_1)\cdot
P_\xi(x_2|x_1)\cdot\ldots\cdot P_\xi(x_n|x^{n-1})$.

A standard Bayesian approach to combine the expert predictions is to
define a prior $w$ on the experts $\Xi$ which induces a joint
distribution with mass function $P(x^n,\xi)=w(\xi)P_\xi(x^n)$.
Inference is then based on this joint distribution. We can compute,
for example: (a) the \emph{marginal probability} of the data
$P(x^n)=\sum_{\xi\in\Xi}w(\xi)P_\xi(x^n)$, (b) the \emph{predictive
 distribution} on the next outcome
$P(\rv{x}_{n+1}|x^n)=P(x^n,\rv{x}_{n+1})/P(x^n)$, which defines a prediction
strategy that combines those of the individual experts, or (c) the
\emph{posterior distribution} on the experts
$P(\rv{\xi}|x^n)=P_{\rv{\xi}}(x^n)w(\rv \xi)/P(x^n)$, which tells us how the
experts' predictions should be weighted. This simple probabilistic
approach has the advantage that it is computationally easy: predicting 
$n$ outcomes using $|\Xi|$ experts requires only $O(n\cdot|\Xi|)$
time. Additionally, this Bayesian strategy guarantees that the overall
probability of the data is only a factor $w(\hat\xi)$ smaller than the
probability of the data according to the best available expert
$\hat\xi$. On the flip side, with this strategy we never do any
\emph{better} than $\hat\xi$ either: we have $P_{\hat\xi}(x^n)\ge
P(x^n)\ge P_{\hat\xi}(x^n)w(\hat\xi)$, which means that potentially
valuable insights from the other experts are not used to our
advantage!

More sophisticated combinations of prediction strategies can be found in the literature under various headings, including (Bayesian) statistics, source coding and universal prediction. In the latter the experts' predictions are not necessarily probabilistic, and scored using an arbitrary loss function. In this paper we consider only logarithmic loss, although our results can undoubtedly be generalised to the framework described in, e.g.\ \cite{Vovk1999}.

We introduce HMMs as an intuitive graphical language that allows unified description of existing and new models. Additionally, the running time for evaluation of such models can be read off directly from the size of their representation.

\subsection*{Overview}
In \secref{sec:es.priors} we develop a more general framework for combining expert
predictions, where we consider the possibility that the \emph{optimal}
weights used to mix the expert predictions may \emph{vary over time}, i.e.\ as the sample size increases. We stick to Bayesian methodology, but we define the
prior distribution as a probability measure on \emph{sequences of
 experts} rather than on experts. The prior probability of a sequence $\xi_1, \xi_2, \ldots$ is the probability that we rely on expert $\xi_1$'s prediction of the first outcome and expert $\xi_2$'s prediction of the second outcome, etc. This allows for the
expression of more sophisticated models for the combination of expert
predictions. For example, the nature of the data generating process
may evolve over time; consequently different experts may be better
during different periods of time. It is also possible that not the
data generating process, but the experts themselves change as more and
more outcomes are being observed: they may learn from past mistakes,
possibly at different rates, or they may have occasional bad days,
etc. In both situations we may hope to benefit from more
sophisticated modelling.

Of course, not all models for combining expert predictions are
computationally feasible. \secref{sec:hmms} describes a methodology
for the specification of models that allow efficient evaluation. We
achieve this by using hidden Markov models (HMMs) on two levels. On
the first level, we use an HMM as a formal specification of a
distribution on sequences of \emph{experts} as defined in
\secref{sec:es.priors}. We introduce a graphical language to
conveniently represent its structure. These graphs help to understand
and compare existing models and to design new ones. We then modify this
first HMM to construct a second HMM that specifies the distribution on
sequences of \emph{outcomes}. Subsequently, we can use the standard
dynamic programming algorithms for HMMs (forward, backward and
Viterbi) on both levels to efficiently calculate most relevant
quantities, most importantly the marginal probability of the observed
outcomes $P(x^n)$ and posterior weights on the next expert given the
previous observations $P(\rv{\xi_{n+1}}|x^n)$.

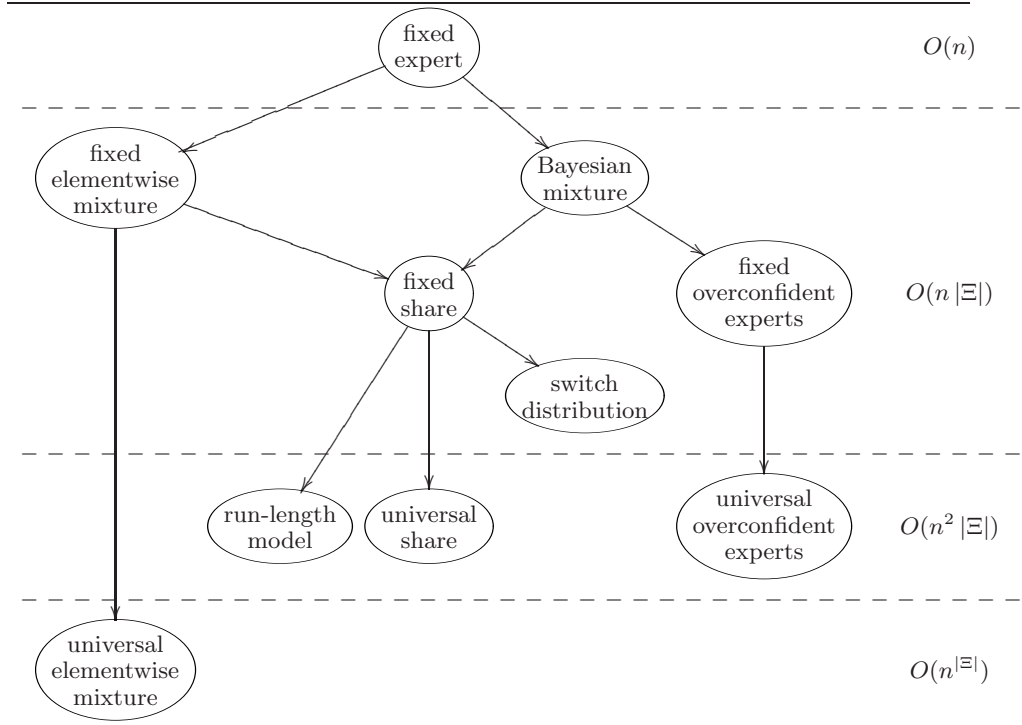
\begin{figure}
\centering
\footnotesize
$\xymatrix@C=0.5em@R=0.5em{
&&& *++[o][F]\txt{fixed\\expert} \ar[ddll] \ar[ddr]&&&&O(n)\\
\ar@{--}[rrrrrrrr]&&&&&&&&\\
&*++[o][F]\txt{fixed\\elementwise\\mixture}\ar[drr]\ar[dddddd]&&&*++[o][F]\txt{Bayesian\\mixture}\ar[dl] \ar[dr]\\
&&&*++[o][F]\txt{fixed\\share}\ar[dr] \ar[dddl] \ar[ddd]&&*++[o][F]\txt{fixed\\overconfident\\experts} \ar[ddd]&& O(n\card{\Xi})\\
&&&& *++[o][F]\txt{switch\\distribution}\\
\ar@{--}[rrrrrrrr]&&&&&&&&\\
&&*++[o][F]\txt{run-length\\model}&*++[o][F]\txt{universal\\share}&  &*++[o][F]\txt{universal\\overconfident\\experts} & & O(n^2\card{\Xi})\\
\ar@{--}[rrrrrrrr]&&&&&&&&\\
&*++[o][F]\txt{universal\\elementwise\\mixture}&&&&&&O(n^{\card{\Xi}})
}$
\caption{Expert sequence priors: generalisation relationships and run time}\label{fig:zoology}
\end{figure}

It turns out that many existing models for prediction with expert advice can be specified as HMMs. We provide an overview in \secref{sec:zoology} by giving the graphical representations of the HMMs corresponding to the following models. First, universal elementwise mixtures (sometimes called mixture models) that learn the optimal mixture parameter from data. Second, Herbster and Warmuth's fixed share algorithm for tracking the best expert \cite{HerbsterWarmuth1995, HerbsterWarmuth1998}. Third, universal share, which was introduced by Volf and Willems as the ``switching method'' \cite{volfwillems1998} and later independently proposed by Bousquet \cite{bousquet2003}. Here the goal is to learn the optimal fixed-share parameter from data. The last considered model safeguards against overconfident experts, a case first considered by Vovk in \cite{Vovk1999}. We render each model as a prior on sequences of experts by giving its HMM. The size of the HMM immediately determines the required running time for the forward algorithm. The generalisation relationships between these models as well as their running times are displayed in \figref{fig:zoology}. In each case this running time coincides with that of the best known algorithm. We also give a loss bound for each model, relating the loss of the model to the loss of the best competitor among a set of alternatives in the worst case. Such loss bounds can help select between different models for specific prediction tasks.

Besides the models found in the literature, \figref{fig:zoology} also
includes two new generalisations of fixed share: the switch
distribution and the run-length model. These models are the subject of
\secref{sec:newstuff}. The switch distribution was introduced
in~\cite{threemusketeers07} as a practical means of improving
Bayes/Minimum Description Length prediction to achieve the optimal rate of convergence in nonparametric settings. Here we give
the concrete HMM that allows for its linear time computation, and we
prove that it matches the parametric definition given
in~\cite{threemusketeers07}. The run-length model is based on a distribution on the number of successive outcomes that are typically well-predicted by the same expert. Run-length codes are typically applied directly to the data, but in our novel application they define the prior on expert sequences instead. Again, we provide the graphical representation of their defining HMMs as well as loss bounds.

Then in \secref{sec:loose.ends} we discuss a number of extensions of
the above approach, such as approximation methods to speed up
calculations for large HMMs.

\section{Expert Sequence Priors}\label{sec:es.priors}
In this section we explain 
how expert tracking can be described in probability theory using expert sequence priors (ES-priors). These ES-priors are distributions on the space of infinite sequences of experts that are used to express regularities in the development of the relative quality of the experts' predictions. As illustrations we render Bayesian mixtures and elementwise mixtures as ES-priors. In the next section we show how ES-priors can be implemented efficiently by hidden Markov models.

\paragraph{Notation}
We denote by $\nats$ the natural numbers including zero, and by $\posints$ the natural numbers excluding zero. For $n \in \nats$, we abbreviate $\set{1,2,\ldots, n}$ by $[n]$. We let $[\omega] := \set{1,2, \ldots}$. Let $\wstates$ be a set. We denote the cardinality of $\wstates$ by $\card{\wstates}$. For any natural number $n$, we let the variable $\wstate^n$ range over the $n$-fold Cartesian product $\wstates^n$, and we write $\wstate^n = \tuple{\wstate_1, \ldots, \wstate_n}$. We also let $\wstate^\omega$ range over $\wstates^\omega$ --- the set of infinite sequences over $\wstates$ --- and write $\wstate^\omega = \tuple{\wstate_1, \ldots}$. We read the statement $\wstate^\lambda \in \wstates^{\le \omega}$ to first bind $\lambda \le \omega$ and subsequently $\wstate^\lambda \in \wstates^\lambda$. If $\wstate^\lambda$ is a sequence, and $\kappa \le \lambda$, then we denote by $\wstate^\kappa$ the prefix of $\wstate^\lambda$ of length $\kappa$. %

\paragraph{Forecasting System}
Let $\xspace$ be a countable outcome space. We use the notation $\xspace^*$ for the set of all finite sequences over $\xspace$ and let $\simplex(\xspace)$ denote the set of all probability mass functions on $\xspace$. A
\emph{(prequential) $\mathcal X$-forecasting system} (PFS) is a function $P:\xspace^*\rightarrow\simplex(\xspace)$ that maps sequences of previous observations to a predictive distribution on the next outcome. Prequential forecasting systems were introduced by Dawid in~\cite{dawid1984}.

\paragraph{Distributions}
We also require probability measures on spaces of infinite sequences. In such a space, a basic event is the set of all continuations of a given prefix. We identify such events with their prefix. Thus a distribution on $\xspace^\omega$ is defined by a function $P:\xspace^*\rightarrow[0,1]$ that satisfies $P(\epsilon)=1$, where
$\epsilon$ is the empty sequence, and for all $n\ge0$, all
$x^n \in \xspace^n$ we have
$\sum_{x\in\xspace}P(x_1,\ldots,x_{n},x)=P(x^n)$. We identify $P$ with the distribution it defines. We write $P(x^n|x^m)$ for
$P(x^n)/P(x^m)$ if $0\le m\le n$. 

Note that forecasting systems continue to
make predictions even after they have assigned probability $0$ to a
previous outcome, while distributions' predictions become undefined. 
Nonetheless we use the same notation: we write $P(x_{n+1}|x^n)$ for the probability that a forecasting system $P$ assigns to the $n+1$st outcome given the first $n$ outcomes, as if $P$ were a
distribution.

\paragraph{ES-Priors} 
The slogan of this paper is \emph{we do not understand the data}. Instead of modelling the data, we work with experts. We assume that there is a fixed set of experts $\Xi$, and that each expert $\xi \in \Xi$ predicts using a forecasting system $P_\xi$. Adopting Bayesian methodology, we impose a prior $\pi$ on infinite sequences of such experts; this prior is called an \markdef{expert sequence prior} (ES-prior). Inference is then based on the distribution on the joint space $\del{\xspace\times\Xi}^\omega$, called the \markdef{ES-joint}, which is defined as follows:
\begin{equation}\label{eq:joint}
\joint\Big(\tuple{\xi_1,x_1},\ldots,\tuple{\xi_n,x_n}\Big):=\esp(\xi^n)\prod_{i=1}^{n}P_{\xi_i}(x_i|x^{i-1}).
\end{equation}
We adopt shorthand notation for events: when we write
$\joint(S)$, where $S$ is a subsequence of $\xi^n$ and/or of $x^n$, this
means the probability under $\joint$ of the set of
sequences of pairs which match $S$ exactly. For example, the marginal probability of a sequence of outcomes is:
\begin{equation}\label{eq:marg}
\joint(x^n)=\sum_{\xi^n\in\Xi^n}\joint(\xi^n,x^n)=\sum_{\xi^n}\joint\Big(\tuple{\xi_1,x_1},\ldots,\tuple{\xi_n,x_n}\Big).
\end{equation}
Compare this to the usual Bayesian statistics, where a model class $\set{P_\theta \mid \theta \in \Theta}$ is also endowed with a prior distribution $w$ on $\Theta$. Then, after observing outcomes $x^n$, inference is based on the posterior $P(\rv \theta| x^n)$ on the parameter, which is never actually observed. Our approach is exactly the same, but we always consider $\Theta = \Xi^\omega$. Thus as usual our predictions are based on the posterior $P(\rv \xi^\omega | x^n)$. However, since the predictive distribution of $\rv{x}_{n+1}$ only depends on $\xi_{n+1}$ (and $x^n$) we always marginalise as follows:
\begin{equation}\label{eq:predictive}
\joint(\xi_{n+1}|x^n)={\joint(\xi_{n+1},x^n)\over\joint(x^n)}={\sum_{\xi^n}\joint(\xi^n,x^n)\cdot\pi(\xi_{n+1}|\xi^n)\over\sum_{\xi^n}\joint(\xi^n,x^n)}.
\end{equation}
At each moment in time we predict the data using the posterior, which is a mixture over our experts' predictions. Ideally, the ES-prior $\pi$ should be chosen such that the posterior coincides with the optimal mixture weights of the experts at each sample size. The traditional interpretation of our ES-prior as a representation of belief about an unknown ``true'' expert sequence is tenuous, as normally the experts do not generate the data, they only predict it. Moreover, by mixing over different expert sequences, it is often possible to predict significantly better than by using any single sequence of experts, a feature that is crucial to the performance of many of the models that will be described below and in \secref{sec:zoology}. In the remainder of this paper we motivate ES-priors by giving performance guarantees in the form of bounds on running time and loss.

\subsection{Examples}\label{sec:es.prior.examples}
We now show how two ubiquitous models can be rendered as ES-priors.

\begin{example}[Bayesian Mixtures]\label{example:bayes.es.prior}
  Let $\Xi$ be a set of experts, and let $P_\xi$ be a PFS for each
  $\xi \in \Xi$. Suppose that we do not know which expert will make
  the best predictions. Following the usual Bayesian methodology, we
  combine their predictions by conceiving a prior $w$ on $\Xi$, which
  (depending on the adhered philosophy) may or may not be interpreted
  as an expression of one's beliefs in this respect. Then the standard
  Bayesian mixture $P_\textnormal{bayes}$ is given by
\begin{equation}\label{eq:bayes}
P_\textnormal{bayes}(x^n) = \sum_{\xi \in \Xi} P_\xi(x^n) w(\xi), \quad\text{where}\quad P_\xi(x^n) = \prod_{i =1}^n P_\xi(x_i|x^i).
\end{equation}
The Bayesian mixture is not an ES-joint, but it can easily be transformed into one by using the ES-prior that assigns probability $w(\xi)$ to the identically-$\xi$ sequence for each $\xi \in \Xi$:
\[ \pi_\textnormal{bayes}(\xi^n) = 
\begin{cases}
w(k) & \text{if $\xi_i = k$ for all $i = 1,\ldots,n$,} \\ 0 & \text{o.w.}
\end{cases}
\]
We will use the adjective ``Bayesian'' generously throughout this
paper, but when we write \markdef{the standard Bayesian ES-prior} this
 always refers to $\pi_\textnormal{bayes}$.
\qedex
\end{example}

\begin{example}[Elementwise Mixtures]\label{example:fixed.elementwise.mixtures}
The \markdef{elementwise mixture}\footnote{These mixtures are sometimes just called mixtures, or predictive mixtures. We use the term elementwise mixtures both for descriptive clarity and to avoid confusion with Bayesian mixtures. } is formed from some mixture weights $\alpha \in \simplex(\Xi)$ by
\[ P_{\text{mix},\alpha}(x^n) := \prod_{i=1}^n P_\alpha(x_i|x^{i-1}), 
\quad\text{where}\quad 
P_\alpha(x_n|x^{n-1}) = \sum_{\xi \in \Xi} P_\xi(x_n|x^{n-1}) \alpha(\xi).\]
In the preceding definition, it may seem that elementwise mixtures do
not fit in the framework of ES-priors. But we can rewrite this
definition in the required form as follows:
\begin{subequations}\label{ex:fix-mix-inward}
\begin{gather}
\begin{split}
P_{\text{mix},\alpha}(x^n)
 &= \prod_{i=1}^n \sum_{\xi \in\Xi} P_\xi(x_i|x^{i-1}) \alpha(\xi) = \sum_{\xi^n \in \Xi^n} \prod_{i=1}^n P_{\xi_i}(x_i|x^{i-1}) \alpha(\xi_i) \\
 &= \sum_{\xi^n} P(x^n|\xi^n) \pi_{\text{mix},\alpha}(\xi^n), 
\end{split}
\intertext{which is the ES-joint based on the prior} \label{eq:pi.mix}
\pi_{\text{mix},\alpha}(\xi^n) := \prod_{i=1}^n\alpha(\xi_i).
\end{gather}
\end{subequations}
Thus, the ES-prior for elementwise mixtures is just the multinomial distribution with mixture weights $\alpha$.
\qedex
\end{example}
\noindent We mentioned above that ES-priors cannot be interpreted as
expressions of belief about individual expert sequences; this is a
prime example where the ES-prior is crafted such that its posterior
$\pi_{\text{mix},\alpha}(\xi_{n+1}|\xi^n)$ exactly coincides with the
desired mixture of experts.

\section{Expert Tracking using HMMs}\label{sec:hmms}
We explained in the previous section how expert tracking can be implemented using expert sequence priors. In this section we specify ES-priors using hidden Markov models (HMMs). The advantage of using HMMs is that the complexity of the resulting expert tracking procedure can be read off directly from the structure of the HMM. We first give a short overview of the particular kind of HMMs that we use throughout this paper. We then show how HMMs can be used to specify ES-priors. As illustrations we render the ES-priors that we obtained for Bayesian mixtures and elementwise mixtures in the previous sections as HMMs. We conclude by giving the forward algorithm for our particular kind of HMMs. In \secref{sec:zoology} we provide an overview of ES-priors and their defining HMMs that are found in the literature.

\subsection{Hidden Markov Models Overview}
Hidden Markov models (HMMs) are a well-known tool for specifying
probability distributions on sequences with temporal
structure. Furthermore, these distributions are very appealing
algorithmically: many important probabilities can be computed
efficiently for HMMs. These properties make HMMs ideal models of
expert sequences: ES-priors. For an introduction to HMMs,
see~\cite{rabiner1989}. We require a slightly more general notion that
incorporates silent states and forecasting systems as explained below.

We define our HMMs on a generic set of outcomes $\mathcal O$ to avoid
confusion in later sections, where we use HMMs in two different
contexts. First in \secref{sec:hmmes}, we use HMMs to define
ES-priors, and instantiate $\mathcal O$ with the set of experts $\Xi$.
Then in \secref{sec:hmmdata} we modify the HMM that defines the
ES-prior to incorporate the experts' predictions, whereupon $\mathcal
O$ is instantiated with the set of observable outcomes $\xspace$.

\begin{definition}
Let $\mathcal O$ be a finite set of outcomes. We call a quintuple
\[ \A = \tuple{\wstates, \wstates_{\prd}, \winit,
  \wtf, \family{P_\wstate}{\wstate \in \wstates_\prd}}\] a
\markdef{hidden Markov model} on $\mathcal O$ if $\wstates$ is a
countable set, $\wstates_\prd \subseteq \wstates$, $\winit
\in \simplex(\wstates)$, $\wtf : \wstates \to \simplex(\wstates)$ and
$P_\wstate$ is an $\mathcal O$-forecasting system for each $\wstate
\in \wstates_\prd$.
\end{definition}

\paragraph{Terminology and Notation} We call the elements of $\wstates$ \markdef{states}. We call the states in $\wstates_\prd$ \markdef{productive} and the other states \markdef{silent}. We call $\winit$ the \markdef{initial distribution}, let $I$ denote its support (i.e.\ $I := \set{ \wstate \in \wstates \mid \winit(\wstate) > 0}$) and call $I$ the set of \markdef{initial states}. We call $\wtf$ the \markdef{stochastic transition function}. We let $S_\wstate$ denote the support of $\wtf(\wstate)$, and call each $\wstate' \in S_\wstate$ a \markdef{direct successor} of $\wstate$. We abbreviate $\wtf(\wstate)(\wstate')$ to $\wtf(\wstate \to \wstate')$.
A finite or infinite sequence of states $\wstate^\lambda \in
\wstates^{\le \omega}$ is called a \markdef{branch} through
$\A$. A branch $\wstate^\lambda$ is called a \markdef{run} if
either $\lambda = 0$ (so $\wstate^\lambda = \epsilon$), or $\wstate_1
\in I$ and $\wstate_{i+1} \in S_{\wstate_i}$ for all $1 \le i <
\lambda$. A finite run $\wstate^n \neq \epsilon$ is called \markdef{%
a run %
to $\wstate_n$}.
For each branch $\wstate^\lambda$, we denote by $\wstate^\lambda_\prd$ its subsequence of productive states. We denote the elements of $\wstate^\lambda_\prd$ by $\wstate^\prd_1$, $\wstate^\prd_2$ etc.
We call an HMM \markdef{continuous} if $\wstate^\omega_\prd$ is infinite for each infinite run $\wstate^\omega$. 

\paragraph{Restriction} In this paper we will only work with continuous HMMs. This restriction is necessary for the following to be well-defined.

\begin{definition}\label{def:HMM.joint}
An HMM $\A$ induces a joint distribution on runs and sequences of outcomes. Let $o^n \in \mathcal O^n$ be a sequence of outcomes and let $\wstate^{\lambda} \neq \epsilon$ be a run with at least $n$ productive states, then
\[ 
P_{\A}(o^n, \wstate^{\lambda}) := 
\winit(\wstate_1) 
\left(\prod_{i=1}^{\lambda-1} \wtf(\wstate_i \to \wstate_{i+1})\right)
\left(\prod_{i=1}^n P_{\wstate^\prd_i}(o_i | o^{i-1})\right).
\]
The value of $P_\A$ at arguments $o^n, \wstate^\lambda$ that do not fulfil the condition above is determined by the additivity axiom of probability.
\end{definition}

\paragraph{Generative Perspective}
The corresponding generative viewpoint is the following. Begin by
sampling an initial state $\wstate_1$ from the initial distribution
$\winit$. Then iteratively sample a direct successor $\wstate_{i+1}$
from $\wtf(\wstate_{i})$. Whenever a productive state $\wstate_i$ is sampled, say
the $n^\text{th}$, also sample an outcome $o_n$ from the
forecasting system $P_{\wstate_i}$ given all previously sampled
outcomes $o^{n-1}$.

\paragraph{The Importance of Silent States}
Silent states can always be eliminated. Let $\wstate'$ be a silent
state and let $R_{\wstate'} := \set{ \wstate \mid \wstate' \in
  S_\wstate}$ be the set of states that have $\wstate'$ as their
direct successor. Now by connecting each state $\wstate \in
R_{\wstate'}$ to each state $\wstate'' \in S_{\wstate'}$ with
transition probability $\wtf(\wstate \to \wstate')\wtf(\wstate' \to
\wstate'')$ and removing $\wstate'$ we preserve the induced
distribution on $\wstates^\omega$. Now if $\card{R_{\wstate'}} = 1$ or
$\card{S_{\wstate'}} = 1$ then $\wstate'$ deserves this treatment.
Otherwise, the number of successors has increased, since $
\card{R_{\wstate'}}\cdot \card{S_{\wstate'}} \ge \card{R_{\wstate'}} +
\card{S_{\wstate'}}$, and the increase is quadratic in the worst case.
Thus, silent states are important to keep our HMMs small: they can be
viewed as shared common subexpressions. It is important to keep HMMs
small, since the size of an HMM is directly related to the running
time of standard algorithms that operate on it. These algorithms are
described in the next section.

\subsubsection{Algorithms}\label{sec:hmm.algs}
There are three classical tasks associated with hidden Markov models
\cite{rabiner1989}. To give the complexity of algorithms for these
tasks we need to specify the input size. Here we consider the case
where $\wstates$ is finite. The infinite case will be covered in
\secref{sec:algo}. Let $m := \card{\wstates}$ be the number of states
and $e := \sum_{\wstate \in \wstates} \card{S_\wstate}$ be the number
of transitions with nonzero probability. The three tasks are:
\begin{enumerate}
\item \label{it:marg.prob}
Computing the marginal probability $P(o^n)$ of the data $o^n$. 
This task is performed by the forward algorithm. This is a dynamic programming algorithm with time complexity $O(ne)$ and space requirement $O(m)$.
\item \label{it:map.est}
MAP estimation: computing a sequence of states $\wstate^\lambda$ with maximal posterior weight $P(\wstate^\lambda | o^n)$. Note that $\lambda \ge n$.
This task is solved using the Viterbi algorithm, again a dynamic programming algorithm with time complexity $O(\lambda e)$ and space complexity $O(\lambda m)$. 
\item \label{it:par.est}
Parameter estimation. Instead of a single probabilistic transition function $\wtf$, one often considers a collection of transition functions $\tuple{\wtf_\theta \mid \theta \in \Theta}$ indexed by a set of parameters $\Theta$. In this case one often wants to find the parameter $\theta$ for which the HMM using transition function $\wtf_\theta$ achieves highest likelihood $P(o^n|\theta)$ of the data $o^n$.

This task is solved using the Baum-Welch algorithm. This is an iterative improvement algorithm (in fact an instance of Expectation Maximisation (EM)) built atop the forward algorithm (and a related dynamic programming algorithm called the backward algorithm).
\end{enumerate}
Since we apply HMMs to sequential prediction, in this paper we are
mainly concerned with Task~\ref{it:marg.prob} and occasionally with
Task~\ref{it:map.est}. Task~\ref{it:par.est} is outside the scope of this
study.

We note that the forward and backward algorithms actually compute more information than just the marginal probability $P(o^n)$. They compute $P(\wstate_i^\prd, o^i)$ (forward) and $P(o^n | \wstate_i^\prd, o^i)$ (backward) for each $i=1,\ldots,n$. The forward algorithm can be computed incrementally, and can thus be used for on-line prediction. Forward-backward can be used together to compute $P(\wstate_i^\prd | o^n)$ for $i=1,\ldots, n$, a useful tool in data analysis.

Finally, we note that these algorithms are defined e.g.\ in~\cite{rabiner1989} for HMMs without silent states and with simple distributions on outcomes instead of forecasting systems. All these algorithms can be adapted straightforwardly to our general case. We formulate the forward algorithm for general HMMs in \secref{sec:algo} as an example.

\subsection{HMMs as ES-Priors}\label{sec:hmmes}
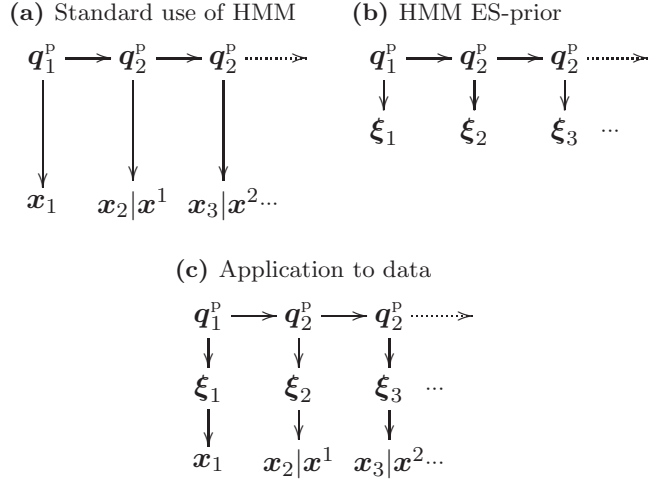
\begin{figure}
\caption{HMMs. $\rv \wstate^\prd_i$, $\rv \xi_i$ and $\rv x_i$ are the $i^\text{th}$ productive state, expert and observation.}
\centering
\subfloat[Standard use of HMM]{\label{fig:hmm}
\patchbl{\xymatrix@R=1em@!C=.9em{
{ \rv \wstate^\prd_1} \ar[r] \ar[dd] & {\rv \wstate^\prd_2} \ar[r] \ar[dd] & {\rv \wstate^\prd_2} \ar@{.>}[r] \ar[dd] & 
\\
{\vphantom{\xi_1}}
\\
{\rv x_1} & {\rv x_2|\rv x^1} & {\rv x_3|\rv x^2} \ar@{}[r]|\cdots & 
}}}
~
\subfloat[HMM ES-prior]{\label{fig:hmm.es.prior}
\patchbl{\xymatrix@R=1em@!C=.9em{
{ \rv \wstate^\prd_1} \ar[r] \ar[d] & {\rv \wstate^\prd_2} \ar[r] \ar[d] & {\rv \wstate^\prd_2} \ar@{.>}[r] \ar[d] & 
\\
{\rv \xi_1} & {\rv \xi_2} & {\rv \xi_3} \ar@{}[r]|\cdots & 
\\
{\vphantom{x_1}}
}}}
~
\subfloat[Application to data]{\label{fig:hmm.prior}
\patchbl{\xymatrix@R=1em@!C=.9em{
{ \rv \wstate^\prd_1} \ar[r] \ar[d] & {\rv \wstate^\prd_2} \ar[r] \ar[d] & {\rv \wstate^\prd_2} \ar@{.>}[r] \ar[d] & 
\\
{\rv \xi_1} \ar[d] & {\rv \xi_2} \ar[d] & {\rv \xi_3} \ar[d]  \ar@{}[r]|\cdots &
\\
{\rv x_1} & {\rv x_2|\rv x^1} & {\rv x_3|\rv x^2} \ar@{}[r]|\cdots &
}}}
\end{figure}

In applications HMMs are often used to model data. This is a good idea whenever there are local temporal correlations between outcomes. A graphical model depicting this approach is displayed in \figref{fig:hmm}.

In this paper we take a different approach; we use HMMs as ES-priors,
that is, to specify temporal correlations between the performance of
our experts. Thus instead of concrete observations our HMMs will
produce sequences of experts, that are never actually observed.
\figref{fig:hmm.es.prior}. illustrates this approach.

Using HMMs as priors allows us to use the standard algorithms of \secref{sec:hmm.algs} to answer questions about the prior. For example, we can use the forward algorithm to compute the prior probability of the sequence of one hundred experts that issues the first expert at all odd time-points and the second expert at all even moments.

Of course, we are often interested in questions about the data rather
than about the prior. In \secref{sec:hmmdata} we show how joints based
on HMM priors (\figref{fig:hmm.prior}) can be transformed into
ordinary HMMs (\figref{fig:hmm}) with at most a $\card{\Xi}$-fold
increase in size, allowing us to use the standard algorithms of
\secref{sec:hmm.algs} not only for the experts, but for the data as
well, with the same increase in complexity. This is the best we can
generally hope for, as we now need to integrate over all possible
expert sequences instead of considering only a single one. Here we
first consider properties of HMMs that represent ES-priors.

\paragraph{Restriction}
HMM priors ``generate'', or define the distribution on, sequences of
experts. But contrary to the data, which are observed, no concrete
sequence of experts is realised. This means that we cannot condition
the distribution on experts in a productive state $\wstate^\prd_n$ on
the sequence of previously produced experts $\xi^{n-1}$. In
other words, we can only use an HMM on $\Xi$ as an ES-prior if the
forecasting systems in its states are simply distributions, so that
all dependencies between consecutive experts are carried by the state.
This is necessary to avoid having to sum over all (exponentially many)
possible expert sequences.

\paragraph{Deterministic} Under the restriction above, but in the presence of silent states, we can make any HMM deterministic in the sense that each forecasting system assigns probability one to a single outcome. We just replace each productive state $\wstate \in \wstates_\prd$ by the following gadget:

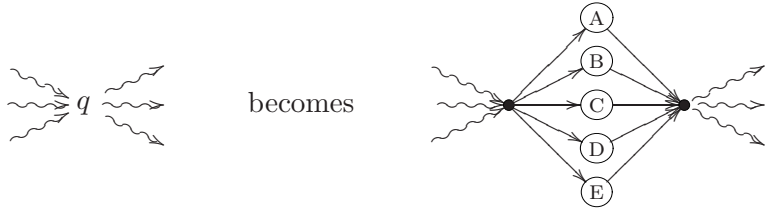
\begin{center}
$\xymatrix@!0@R=1.5em@C=3em{
\\
\ar@{~>}[dr] &&\\
\ar@{~>}[r]  & \wstate \ar@{~>}[r] \ar@{~>}[ur] \ar@{~>}[dr] & \\
\ar@{~>}[ur] && \\
}$
\qquad\raisebox{-3em}{becomes}\qquad
$\xymatrix@!0@R=1.5em@C=3em{
&& \ex{a} \ar[ddr] 
\\
\ar@{~>}[dr] && \ex{b} \ar[dr] &&
\\
\ar@{~>}[r] &\si \ar[uur] \ar[ur] \ar[r] \ar[dr] \ar[ddr] & \ex{c}  \ar[r] & \si \ar@{~>}[r] \ar@{~>}[ur] \ar@{~>}[dr] &
\\
\ar@{~>}[ur] && \ex{d} \ar[ur] &&
\\
&& \ex{e} \ar[uur] 
}$
\end{center}
In the left diagram, the state $\wstate$ has distribution $P_\wstate$
on outcomes $\mathcal O = \set{\expname a,\ldots,\expname e}$. In the
right diagram, the leftmost silent state has transition probability
$P_\wstate(o)$ to a state that deterministically outputs outcome
$o$. We often make the functional relationship explicit and call
$\tuple{\wstates, \wstates_\prd, \winit, \wtf, \wnl}$ a
\markdef{deterministic HMM} on $\mathcal O$ if $\wnl : \wstates_\prd
\to \mathcal O$. Here we slightly abuse notation; the last component of a (general) HMM assigns a \emph{PFS} to each productive state, while the last component of a deterministic HMM  assigns an \emph{outcome} to each productive states.

Sequential prediction using a general HMM or its deterministic
counterpart costs the same amount of work: the $\card{\mathcal O}$-fold increase in the number of states is compensated by the
$\card{\mathcal O}$-fold reduction in the number of outcomes that need
to be considered per state.

\paragraph{Diagrams} Deterministic HMMs can be graphically represented by pictures. In general, we draw a node $N_\wstate$ for each state $\wstate$. We draw a small black dot, e.g.\ $\xymatrix{\si}$, for a silent state, and an ellipse labelled $\wnl(q)$, e.g.\ $\xymatrix{\ex{d}}$, for a productive state. We draw an arrow from $N_\wstate$ to $N_{\wstate'}$ if $\wstate'$ is a direct successor of $\wstate$. We often reify the initial distribution $\winit$ by including a virtual node, drawn as an open circle, e.g.\ $\xymatrix{\bn}$, with an outgoing arrow to $N_\wstate$ for each initial state $\wstate \in I$. The transition probability $P(\wstate \to \wstate')$ is not displayed in the graph.

\subsection{Examples}\label{sec:examples.hmm}
We are now ready to give the deterministic HMMs that correspond to the ES-priors of our earlier examples from \secref{sec:es.prior.examples}: Bayesian mixtures and elementwise mixtures with fixed parameters.

\begin{example}[HMM for Bayesian Mixtures]\label{example:bayes.hmm}
The Bayesian mixture ES-prior $\pi_\textnormal{bayes}$ as introduced in \exref{example:bayes.es.prior} represents the hypothesis that a single expert predicts best for all sample sizes. A simple deterministic HMM that generates the prior $\pi_\textnormal{bayes}$ is given by $\A_{\textnormal{bayes}} = \tuple{\wstates, \wstates_\prd, \wtf, \winit, \Xi, \wnl}$, where
\begin{subequations}\label{eq:esabayes}
\begin{align}
  \wstates = \wstates_\prd &=  \Xi \times \posints 
& \wtf\del{\tuple{\xi, n} \to \tuple{\xi, n+1}} &= 1 
\\
\wnl(\xi,n) &= \xi
& \winit\del{\xi,1} &= w(\xi)
\end{align}
\end{subequations}
The diagram of~\eqref{eq:esabayes} is displayed in \figref{graph:bayes}. 
\begin{figure}
\centering
$\xymatrix@R=0.7em{
&\ex{a}\name{\expname a,1} \ar[r] & \ex{a} \name{\expname a,2} \ar[r] & \ex{a} \ar[r] & \ex{a} \ar@{.>}[r] & \\
\\
&\ex{b} \name{\expname b,1} \ar[r] & \ex{b} \name{\expname b,2} \ar[r] & \ex{b} \ar[r] & \ex{b} \ar@{.>}[r] & \\
\bn  \ar[ur] \ar[uuur] \ar[dr] \ar[dddr]\\
&\ex{c} \name{\expname c,1} \ar[r] & \ex{c} \name{\expname c,2} \ar[r] & \ex{c} \ar[r] & \ex{c} \ar@{.>}[r] & \\
\\
&\ex{d} \name[+<4pt,10pt>]{\expname d,1} \ar[r] & \ex{d} \name{\expname d,2}\ar[r] & \ex{d} \ar[r] & \ex{d} \ar@{.>}[r] & \\
}$
\caption{Combination of four experts using a standard Bayesian mixture.}\label{graph:bayes}
\end{figure}
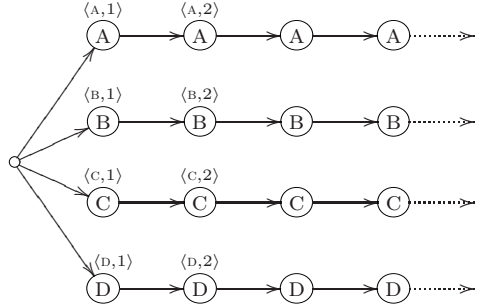
From the picture of the HMM it is clear that it computes the Bayesian mixture. Hence, using~\eqref{eq:bayes}, the loss of the HMM with prior $w$ is bounded for all $x^n$ by
\begin{equation} \label{eq:bayes.loss.bound}
- \log P_{\A_{\textnormal{bayes}}}(x^n) + \log P_\xi(x^n) \le - \log w(\xi) \qquad \text{for all experts $\xi$.}
\end{equation}
In particular this bound holds for $\hat \xi = \argmax_\xi P_\xi(x^n)$, so we predict as well as the  single best expert with \emph{constant} overhead. Also $P_{\A_{\textnormal{bayes}}}(x^n)$ can obviously be computed in $O(n\card{\Xi})$ using its definition~\eqref{eq:bayes}. We show in \secref{sec:algo} that computing it using the HMM prior above gives the same running time $O(n\card{\Xi})$, a perfect match.
\qedex
\end{example}

\begin{example}[HMM for Elementwise Mixtures] \label{example:fixed.elementwise.mixtures.hmm}
We now present the deterministic HMM $\A_{\text{mix},\alpha}$ that implements the ES-prior $\pi_{\text{mix},\alpha}$ of \exref{example:fixed.elementwise.mixtures}. Its diagram is displayed in \figref{graph:fixmix}. The HMM has a single silent state per outcome, and its transition probabilities are the mixture weights $\alpha$. Formally, $\A_{\text{mix},\alpha}$ is given using $\wstates = \wstates_\sil \cup \wstates_\prd$ by
\begin{subequations}
\begin{gather}
\begin{aligned}
\wstates_\sil  &=  \set{\textsf{p}} \times \nats
& \wstates_\prd &= \Xi \times \posints
&
\winit(\textsf{p},0) &= 1
&
\wnl(\xi, n) &= \xi 
\end{aligned}
\\
\wtf\del{
\begin{aligned}
\tuple{\textsf{p},n} &\to \tuple{\xi, n+1}
\\
\tuple{\xi,n} &\to \tuple{\textsf{p},n}
\end{aligned}} = \del{
\begin{gathered}
\alpha(\xi)
\\
1
\end{gathered}}
\end{gather}
\end{subequations}
The vector-style definition of $\wtf$ is shorthand for one $\wtf$ per line.
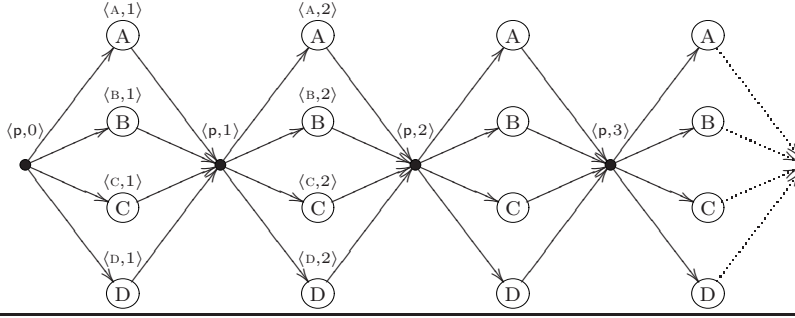
\begin{figure}
\centering
$\xymatrix@R=0.7em@C=2.5em{
&\ex{a} \name{\expname a,1} \ar[rddd] && \ex{a} \name{\expname a,2} \ar[rddd] && \ex{a} \ar[rddd] && \ex{a} \ar@{.>}[rddd] & \\
\\
&\ex{b} \name{\expname b,1} \ar[rd] && \ex{b} \name{\expname b,2} \ar[rd] && \ex{b} \ar[rd] && \ex{b} \ar@{.>}[rd] & \\
\si \name[+<0pt,13pt>]{\textsf{p},0} \ar[ur] \ar[uuur] \ar[dr] \ar[dddr]&&
\si \name[+<0pt,13pt>]{\textsf{p},1} \ar[ur] \ar[uuur] \ar[dr] \ar[dddr]&&
\si \name[+<0pt,13pt>]{\textsf{p},2} \ar[ur] \ar[uuur] \ar[dr] \ar[dddr]&&
\si \name[+<0pt,13pt>]{\textsf{p},3} \ar[ur] \ar[uuur] \ar[dr] \ar[dddr]&&
\\
&\ex{c} \name{\expname c,1} \ar[ru] && \ex{c} \name{\expname c,2} \ar[ru] && \ex{c} \ar[ru] && \ex{c} \ar@{.>}[ru] & \\
\\
&\ex{d} \name[+<0pt,13pt>]{\expname d,1} \ar[ruuu] && \ex{d} \name[+<0pt,13pt>]{\expname d,2} \ar[ruuu] && \ex{d} \ar[ruuu] && \ex{d} \ar@{.>}[ruuu] &
}$
\caption{Combination of four experts using a fixed elementwise mixture}\label{graph:fixmix}
\end{figure}%
We show in \secref{sec:algo} that this HMM allows us to compute $P_{\A_{\text{mix},\alpha}}(x^n)$ in time $O(n\card{\Xi})$.
\qedex
\end{example}

\subsection{The HMM for Data}\label{sec:hmmdata}
We obtain our model for the data (\figref{fig:hmm.prior}) by composing an HMM prior on $\Xi^\omega$ with a PFS $P_\xi$ for each expert $\xi \in \Xi$. We now show that the resulting marginal distribution on data can be implemented by a single HMM on $\xspace$ (\figref{fig:hmm}) \emph{with the same number of states as the HMM prior}. Let $P_\xi$ be an $\xspace$-forecasting system for each $\xi \in \Xi$, and let the ES-prior $\pi_{\A}$ be given by the deterministic HMM $\A = \tuple{\wstates, \wstates_\prd, \winit, \wtf, \wnl}$ on $\Xi$. Then the marginal distribution of the data (see \eqref{eq:joint}) is given by \[P_{\A}(x^n) = \sum_{\xi^n} \pi_{\A}(\xi^n) \prod_{i=1}^n P_{\xi_i}(x_i|x^{i-1}).\]
The HMM
$\mathbb X := \tuple{\wstates, \wstates_\prd, \winit, \wtf, \family{P_{\wnl(\wstate)}}{\wstate \in \wstates_\prd}}$ on $\xspace$ induces the same marginal distribution (see \defref{def:HMM.joint}). That is, $P_{\mathbb X}(x^n) = P_{\A}(x^n)$. Moreover, $\mathbb X$ contains only the forecasting systems that also exist in $\A$ and it retains the structure of $\A$. In particular this means that the HMM algorithms of \secref{sec:hmm.algs} have the \emph{same} running time on the prior $\A$ as on the marginal $\mathbb X$.

\subsection{The Forward Algorithm and Sequential Prediction}\label{sec:algo}
We claimed in \secref{sec:hmm.algs} that the standard HMM algorithms could easily be extended to our HMMs with silent states and forecasting systems. In this section we give the main example: the forward algorithm. We will also show how it can be applied to sequential prediction. Recall that the forward algorithm computes the marginal probability $P(x^n)$ for fixed $x^n$. On the other hand, sequential prediction means predicting the next \emph{observation} $\rv{x}_{n+1}$ for given data $x^n$, i.e.\ computing its distribution. For this it suffices to predict the next \emph{expert} $\rv{\xi}_{n+1}$; we then simply predict $\rv{x}_{n+1}$ by averaging the expert's predictions accordingly: $P(x_{n+1}|x^n) = \Exp[{P_{\rv{\xi}_{n+1}}(x_{n+1}|x^n)}]$.

We first describe the preprocessing step called \emph{unfolding} and introduce notation for nodes. We then give the forward algorithm, prove its correctness and analyse its running time and space requirement. The forward algorithm can be used for prediction with expert advice. We conclude by outlining the difficulty of adapting the Viterbi algorithm for MAP estimation to the expert setting.

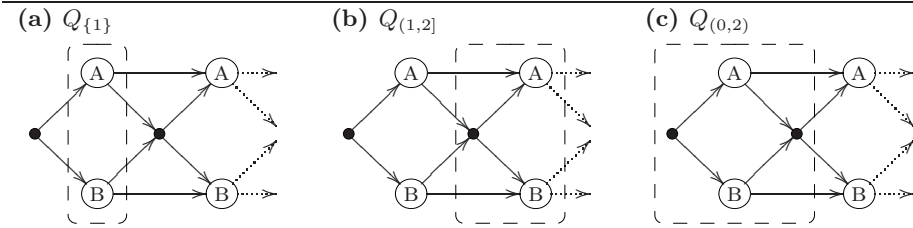
\begin{figure}
\caption{Unfolding example}
\centering
\begin{minipage}{0.3\textwidth}
\subfloat[Prior to unfolding]{\label{fig:pre.unfolding}
\patchbl{\xymatrix{
\quad & \ex{a} \ar@(dr,dl)[] & \quad
}}}%
\\
\subfloat[After unfolding]{\label{fig:post.unfolding}
\patchbl{\xymatrix@C=1em{
\ex{a} \ar[r] & \ex{a} \ar[r] & \ex{a} \ar@{.>}[r] &
}}
}%
\end{minipage}%
\begin{minipage}{0.3\textwidth}
\subfloat[Bayesian mixture]{\label{fig:bayes.folded}
\qquad $\xymatrix@R=0.7em{
&\ex{a}\ar@(ur,dr)[]
\\
\\
&\ex{b} \ar@(ur,dr)[]
\\
\bn  \ar[ur] \ar[uuur] \ar[dr] \ar[dddr]
\\
&\ex{c} \ar@(ur,dr)[]
\\
\\
&\ex{d}\ar@(ur,dr)[]
}$\qquad
}
\end{minipage}
\begin{minipage}{0.3\textwidth}
\subfloat[Elementwise mixture]{\label{fig:eltwise.mixtures.folded}
~\qquad$\xymatrix@R=0.7em@C=2.5em{
&\ex{a} 
\\
\\
&\ex{b} 
\\
\si \ar@{<->}[ur] \ar@{<->}[uuur] \ar@{<->}[dr] \ar@{<->}[dddr]&
\\
&\ex{c} 
\\
\\
&\ex{d} 
}$\qquad~
}
\end{minipage}
\end{figure}

\paragraph{Unfolding} Every HMM can be transformed into an equivalent HMM in which each productive state is involved in the production of a unique outcome. The single node in \figref{fig:pre.unfolding} is involved in the production of $\rv x_1, \rv x_2, \ldots$ In its unfolding \figref{fig:post.unfolding} the $i^\text{th}$ node is only involved in producing $\rv x_i$. Figures~\ref{fig:bayes.folded} and \ref{fig:eltwise.mixtures.folded} show HMMs that unfold to the Bayesian mixture shown in \figref{graph:bayes} and the elementwise mixture shown in \figref{graph:fixmix}. In full generality, fix an HMM $\A$. The \markdef{unfolding} of $\A$ is the HMM
\[ \A^\unf := \tuple{\wstates^\unf, \wstates^\unf_\prd, \winit^\unf, \wtf^\unf, \family{P^\unf_\wstate}{\wstate \in \wstates^\unf}},\]
where the states and productive states are given by:
\begin{subequations}
\begin{align}
\wstates^\unf &:= \set{\tuple{\wstate_\lambda, n} \mid \text{$\wstate^\lambda$ is a run through $\A$}},\quad \text{where $n=\card{\wstate^\lambda_\prd}$}
\\
\wstates^\unf_\prd &:= \wstates^\unf \cap (\wstates_\prd \times \nats)
\end{align}
and the initial probability, transition function and forecasting systems are:
\begin{align}
\winit^\unf\del{\tuple{\wstate, 0}} &:= \winit(\wstate) 
\\
\wtf^\unf\del{\begin{aligned}
\tuple{\wstate, n} &\to \tuple{\wstate',n+1}
\\
\tuple{\wstate, n} &\to \tuple{\wstate',n}
\end{aligned}}
&:=
\del{\begin{gathered}
\wtf(\wstate \to \wstate')
\\
\wtf(\wstate \to \wstate') 
\end{gathered}}
\\
P^\unf_{\tuple{\wstate, n}}  &:= P_\wstate
\end{align}
\end{subequations}
First observe that unfolding preserves the marginal: $P_{\A}(o^n) = P_{\A^\unf}(o^n)$. Second, unfolding is an idempotent operation: $\del{\A^\unf}^\unf$ is isomorphic to $\A^\unf$. Third, unfolding renders the set of states infinite, but for each $n$ it preserves the number of states reachable in exactly $n$ steps.

\paragraph{Order} The states in an unfolded HMM have earlier-later structure. Fix $\wstate, \wstate' \in \wstates^\unf$. We write
$\wstate < \wstate'$ iff there is a run to $\wstate'$ through $\wstate$. We call $<$ the \markdef{natural order} on $\wstates^\unf$. Obviously $<$ is a partial order, furthermore it is the transitive closure of the reverse direct successor relation. It is well-founded, allowing us to perform induction on states, an essential ingredient in the forward algorithm (\algref{alg:forward}) and its correctness proof (\thmref{thm:correctness.of.forward.algorithm}).

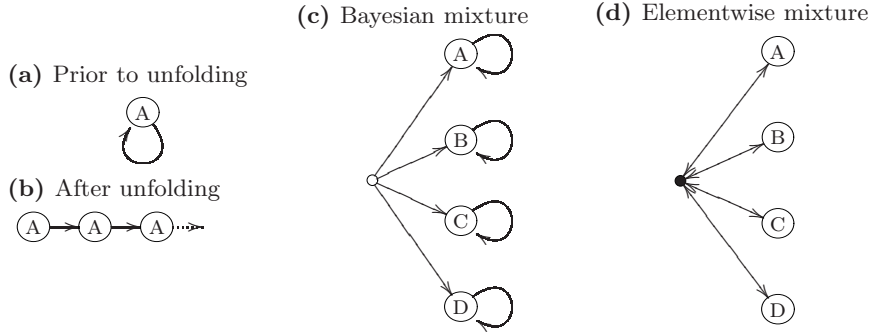
\begin{figure}
\caption{Interval notation}\label{fig:strata}
\centering
\subfloat[$Q_{\set{1}}$]{
$\xymatrix@R=1.3em@C=1.3em{
&          \ex{a} \ar[dr] \ar[rr] && \ex{a} \ar@{..>}[r] \ar@{..>}[dr]&
\\
\si \ar[ur] \ar[dr]  && \si \ar[ur] \ar[dr] &&
\\
&          \ex{b} \ar[ur] \ar[rr]& & \ex{b} \ar@{..>}[r] \ar@{..>}[ur]&
\save"1,2"+0.{"3,2"+0}!C*+<2em>[F--:<3pt>]\frm{}\restore
}$
}
~
\subfloat[$Q_{\intoc{1}{2}}$]{
$\xymatrix@R=1.3em@C=1.3em{
&          \ex{a} \ar[dr] \ar[rr] && \ex{a} \ar@{..>}[r] \ar@{..>}[dr]&
\\
\si \ar[ur] \ar[dr]  && \si \ar[ur] \ar[dr] &&
\\
&          \ex{b} \ar[ur] \ar[rr]&& \ex{b} \ar@{..>}[r] \ar@{..>}[ur]&
\save"1,3"+<0.4em,0em>.{"3,4"+0}!C*+<2em>[F--:<3pt>]\frm{}\restore
}$
}
~
\subfloat[$Q_{\intoo{0}{2}}$]{
$\xymatrix@R=1.3em@C=1.3em{
 & \ex{a} \ar[dr] \ar[rr] && \ex{a} \ar@{..>}[r] \ar@{..>}[dr]&
\\
\si \ar[ur] \ar[dr]  && \si \ar[ur] \ar[dr] &&
\\
 & \ex{b} \ar[ur] \ar[rr]&& \ex{b} \ar@{..>}[r] \ar@{..>}[ur]&
\save"1,1"+<0.4em,0em>.{"3,3"-<0.4em,0em>}!C*+<2em>[F--:<3pt>]\frm{}\restore
}$
}
\end{figure}

\paragraph{Interval Notation} We introduce interval notation to address subsets of states of unfolded HMMs, as illustrated by \figref{fig:strata}.. Our notation associates each productive state with the sample size at which it produces its outcome, while the silent states fall in between. We use intervals with borders in $\nats$. The interval contains the border $i \in \nats$ iff the addressed set of states includes the states where the $i^\text{th}$ observation is produced. 
\begin{subequations}
\begin{align}
\wstates^\unf_{\intco{n}{m}} &:= \wstates^\unf \cap (\wstates \times \intco{n}{m})
&
\wstates^\unf_{\intcc{n}{m}} &:= \wstates^\unf_{\intco{n}{m}} \cup \wstates^\unf_{\set{m}}
\\
\wstates^\unf_{\set{n}} &:= \wstates^\unf \cap (\wstates_\prd \times \set{n})
&
\wstates^\unf_{\intoo{n}{m}} &:= \wstates^\unf_{\intco{n}{m}} \setminus \wstates^\unf_{\set{n}}
\\
& &
\wstates^\unf_{\intoc{n}{m}} &:= \wstates^\unf_{\intcc{n}{m}} \setminus \wstates^\unf_{\set{n}}
\end{align}
\end{subequations}
Fix $n>0$, then $\wstates^\unf_{\set{n}}$ is a non-empty $<$-anti-chain (i.e.\ its states are pairwise $<$-incomparable). Furthermore $\wstates^\unf_{\intoo{n}{n+1}}$ is empty iff $\wstates^\unf_{\set{n+1}} = \bigcup_{\wstate \in \wstates^\unf_{\set{n}}} S_\wstate$, in other words, if there are no silent states between sample sizes $n$ and $n+1$. 

\paragraph{The Forward Algorithm} The forward algorithm is shown as \algref{alg:forward}. 

\begin{algorithm}
\caption{Concurrent Forward Algorithm and Sequential Prediction. Fix an unfolded deterministic HMM prior $\A = \tuple{\wstates, \wstates_\prd, \winit, \wtf, \wnl}$ on $\Xi$, and an $\xspace$-PFS $P_\xi$ for each expert $\xi \in \Xi$. The input consists of a sequence $x^\omega$ that arrives sequentially.}\label{alg:forward}
\begin{algorithmic}
\STATE Declare the weight map (partial function) $w \partialfn \wstates \to \intcc{0}{1}$.
\STATE $w(v) \gets \winit(v)$ \textbf{for all} $v$ s.t.\ $\winit(v) > 0$. \COMMENT{$\dom(w) = I$}
\FOR{$n=1,2,\ldots$}
\STATE \textsc{Forward Propagation($n$)}
\STATE Predict next expert: $P(\rv{\xi}_n = \xi|x^{n-1}) = \raisebox{0pt}[0pt][0pt]{$\dfrac{\sum_{v \in \wstates_{\set{n}} : \wnl(v) = \xi} w(v)}{\sum_{v \in \wstates_{\set{n}}} w(v)}$}$.
\STATE \textsc{Loss Update($n$)}
\STATE Report probability of data: $P(x^n) = \sum_{v \in \wstates_{\set{n}}} w(v)$.
\ENDFOR
\bigskip
\begin{procedure}{\textsc{Forward Propagation($n$)}}
\WHILE[$\dom(w) \subseteq \wstates_{\intcc{n-1}{n}}$]{$\dom(w) \neq \wstates_{\set{n}}$}
\STATE Pick a $<$-minimal state $u$ in $\dom(w) \setminus \wstates_{\set{n}}$. \COMMENT{$u \in Q_{\intco{n-1}{n}}$}
\FORALL[$v \in Q_{\intoc{n-1}{n}}$]{$v \in S_u$}
  \STATE $w(v) \gets 0$ \textbf{if} $v \notin \dom(w)$.
  \STATE $w(v) \gets w(v) + w(u) \wtf(u \to v)$. 
\ENDFOR
\STATE Remove $u$ from the domain of $w$.
\ENDWHILE \COMMENT{$\dom(w) = \wstates_{\set{n}}$}
\end{procedure}
\bigskip
\begin{procedure}{\textsc{Loss Update($n$)}}
\FORALL[$v \in \wstates_\prd$]{$v \in \wstates_{\set{n}}$}
\STATE $\displaystyle w(v) \gets w(v) P_{\wnl(v)}(x_n|x^{n-1})$.
\ENDFOR
\end{procedure}
\end{algorithmic}
\end{algorithm}

\paragraph{Analysis} Consider a state $\wstate \in \wstates$, say $\wstate \in Q_{\intco{n}{n+1}}$. Initially, $\wstate \notin \dom(w)$. Then at some point $w(\wstate) \gets \winit(\wstate)$. This happens either in the second line because $\wstate \in I$ or in \textsc{Forward Propagation} because $\wstate \in S_u$ for some $u$ (in this case $\winit(\wstate) = 0$). Then $w(\wstate)$ accumulates weight as its direct predecessors are processed in \textsc{Forward Propagation}. At some point all its predecessors have been processed. If $\wstate$ is productive we call its weight at this point (that is, just before \textsc{Loss Update}) $\Alg(\A, x^{n-1}, \wstate)$. Finally, \textsc{Forward Propagation} removes $\wstate$ from the domain of $w$, never to be considered again. We call the weight of $\wstate$ (silent or productive) just before removal $\Alg(\A, x^n, \wstate)$.

Note that we associate \emph{two} weights with each productive state $\wstate \in \wstates_{\set{n}}$: the weight $\Alg(\A, x^{n-1}, \wstate)$ is calculated \emph{before} outcome $n$ is observed, while $\Alg(\A, x^n, \wstate)$ denotes the weight \emph{after} the loss update incorporates outcome $n$.

\begin{theorem}\label{thm:correctness.of.forward.algorithm}
Fix an HMM prior $\A$, $n \in \nats$ and $\wstate \in \wstates_{\intcc{n}{n+1}}$, then
\[ \Alg(\A, x^n, \wstate) = P_\A(x^n, \wstate).\]
\end{theorem}
\noindent
Note that the theorem applies twice to productive states.
\begin{proof} By $<$-induction on states. Let $\wstate \in \wstates_{\intoc{n}{n+1}}$, and suppose that the theorem holds for all $\wstate' < \wstate$. Let $B_{\wstate} = \set{\wstate' \mid \wtf(\wstate' \to \wstate) > 0}$ be the set of direct predecessors of $\wstate$. Observe that $B_\wstate \subseteq Q_{\intco{n}{n+1}}$. The weight that is accumulated by \textsc{Forward Propagation($n$)} onto $\wstate$ is:
\begin{align*}
 \Alg(\A, x^n, \wstate) &= \winit(\wstate) + \sum_{\wstate' \in B_{\wstate}} \wtf(\wstate' \to \wstate) \Alg(\A,x^n,\wstate') 
\\
&= \winit(\wstate) + \sum_{\wstate' \in B_{\wstate}} \wtf(\wstate' \to \wstate) P_\A(x^n, \wstate')  ~=~ P_\A(x^n, \wstate).
\end{align*}
The second equality follows from the induction hypothesis. Additionally if $\wstate \in Q_{\set{n}}$ is productive, say $\wnl(\wstate) = \xi$, then after \textsc{Loss Update($n$)} its weight is:
\begin{align*}
\Alg(\A, x^{n}, \wstate) 
&= P_\xi(x_{n}|x^{n-1})  \Alg(\A, x^{n-1}, \wstate) 
\\
&= P_\xi(x_{n}|x^{n-1}) P_\A(x^{n-1}, \wstate)
~=~ P_\A(x^{n}, \wstate). \qedhere
\end{align*}
The second inequality holds by induction on $n$, and the third by \defref{def:HMM.joint}.
\end{proof}

\paragraph{Complexity}
We are now able to sharpen the complexity results as listed in
\secref{sec:hmm.algs}, and extend them to infinite HMMs. Fix $\A$, $n \in \nats$. The forward algorithm processes each state in $Q_{\intco{0}{n}}$ once, and at that point this state's weight is distributed over its successors. Thus, the running time is proportional to $\sum_{\wstate \in Q_{\intco{0}{n}}} \card{S_\wstate}$. The forward algorithm keeps $\card{\dom(w)}$ many weights. But at each sample size $n$, $\dom(w) \subseteq Q_{\intcc{n}{n+1}}$. Therefore the space needed is at most proportional to $\max_{m < n} \card{\wstates_{\intcc{m}{m+1}}}$. For both Bayes (\exref{example:bayes.hmm}) and elementwise mixtures (\exref{example:fixed.elementwise.mixtures.hmm}) one may read from the figures that $\sum_{\wstate \in \wstates_{\intco{n}{n+1}}} \card{S_\wstate}$ and $\card{\wstates_{\intco{n}{n+1}}}$ are $O(\card{\Xi})$, so we indeed get the claimed running time $O(n\card{\Xi})$ and space requirement $O(\card{\Xi})$.

\paragraph{MAP Estimation}
The forward algorithm described above computes the probability of the data, that is
\[
P(x^n) 
~=~
\sum_{\wstate^\lambda: \wstate_\lambda \in Q_{\set{n}}} P(x^n,\wstate^\lambda).
\]
Instead of the entire sum, we are sometimes interested in the sequence of states $\wstate^\lambda$ that contributes most to it:
\[
\argmax_{\wstate^\lambda} P(x^n, \wstate^\lambda) ~=~ \argmax_{\wstate^\lambda} P(x^n|\wstate^\lambda) \pi(\wstate^\lambda).
\]
The Viterbi algorithm~\cite{rabiner1989} is used to compute the most likely sequence of states for HMMs. It can be easily adapted to handle silent states. However, we may also write
\[
P(x^n)
~=~
\sum_{\xi^n} P(x^n,\xi^n),
\]
and wonder about the sequence of experts $\xi^n$ that contributes most. This problem is harder because in general, a single sequence of experts can be generated by many different sequences of states. This is unrelated to the presence of the silent states, but due to different states producing the same expert simultaneously (i.e. in the same $Q_{\set{n}}$). So we cannot use the Viterbi algorithm as it is. The Viterbi algorithm can be extended to compute the MAP expert sequence for general HMMs, but the resulting running time explodes.
Still, the MAP $\xi^n$ can be sometimes be obtained efficiently by exploiting the structure of the HMM at hand. The first example is the unambiguous HMMs. A deterministic HMM is \markdef{ambiguous} if it has two runs that agree on the sequence of \emph{experts} produced, but not on the sequence of \emph{productive states}. The straightforward extension of the Viterbi algorithm works for unambiguous HMMs. The second important example is the (ambiguous) switch HMM that we introduce in \secref{sec:switch}. We show how to compute its MAP expert sequence in \secref{sec:switch.map}.

\section{Zoology}\label{sec:zoology}
Perhaps the simplest way to predict using a number of experts is to
pick one of them and mirror her predictions exactly. Beyond this ``fixed
expert model'', we
have considered two methods of combining experts so far, namely taking
Bayesian mixtures, and taking elementwise mixtures as described in
\secref{sec:examples.hmm}. \figref{fig:zoology} shows these
and a number of other, more sophisticated methods that fit in our
framework. The arrows indicate which methods are generalised by which other methods. They have been
partitioned in groups that can be computed in the same amount of time
using HMMs. 

We have presented two examples so far, the Bayesian mixture and the elementwise mixture with fixed coefficients (Examples~\ref{example:bayes.hmm} and \ref{example:fixed.elementwise.mixtures.hmm}). The latter model is parameterised. Choosing a fixed value for the parameter beforehand is often difficult. The first model we discuss learns the optimal parameter value on-line, at the cost of only a small additional loss. We then proceed to discuss a number of important existing expert models.

\subsection{Universal Elementwise Mixtures}\label{sec:mixtures}
A distribution is ``universal'' for a family of distributions if it
incurs small additional loss compared to the best member of the
family.
A standard Bayesian mixture constitutes the simplest example. It is universal for the fixed expert model, where the unknown parameter is the used expert. In \eqref{eq:bayes.loss.bound} we showed that the additional loss is at most $\log \card{\Xi}$ for the uniform prior. 

In \exref{example:fixed.elementwise.mixtures.hmm} we described elementwise mixtures with
fixed coefficients as ES-priors. Prior knowledge about the mixture coefficients is often unavailable. We now expand this model to learn the
optimal mixture coefficients from the data. To this end we place a
prior distribution $w$ on the space of mixture weights $\simplex(\Xi)$. Using
\eqref{ex:fix-mix-inward} we obtain the following marginal
distribution:
\begin{multline}\label{eq:umix}
P_\text{umix}(x^n)
=\int_{\simplex(\Xi)}\!\!\!\!P_{\text{mix},\alpha}(x^n)w(\alpha)\dif\alpha=\int_{\simplex(\Xi)}\sum_{\xi^n}P(x^n|\xi^n)\pi_{\text{mix},\alpha}(\xi^n)w(\alpha)\dif\alpha\\
=\sum_{\xi^n}P(x^n|\xi^n)\pi_\text{umix}(\xi^n),
\quad
\text{where}
\quad
\pi_\text{umix}(\xi^n)=\int_{\simplex(\Xi)}\!\!\!\!\pi_{\text{mix},\alpha}(\xi^n)w(\alpha)\dif\alpha.
\end{multline}
Thus $P_\text{umix}$ is the ES-joint with ES-prior
$\pi_\text{umix}$. This applies more generally: parameters $\alpha$
can be integrated out of an ES-prior regardless of which experts are
used, since the expert predictions $P(x^n|\xi^n)$ do not depend on
$\alpha$.

We will proceed to calculate a loss bound for the universal
elementwise mixture model, showing that it really is universal. After
that we will describe how it can be implemented as a HMM.

\subsubsection{A Loss Bound}
\begin{theorem}\label{thm:unimix}
  Suppose the universal elementwise mixture model is defined using the
  $({1\over2},\ldots,{1\over2})$-Dirichlet prior (that is, Jeffreys'
  prior). Further, let $\hat L=\min_\alpha -\log
  P_{\text{mix},\alpha}(x^n)$ be loss of the fixed elementwise mixture
  weights with maximum likelihood parameter $\alpha$. The additional
  loss incurred by the universal elementwise mixture is bounded by
  \[ -\log P_\textnormal{umix}(x^n)-\hat L\quad\le\quad{\card{\Xi}-1\over2}\log{n\over\pi}+c\]
  for a fixed constant $c$.
\end{theorem}
\noindent To prove this, we first establish the following lemma.

\begin{lemma}\label{lemma:transferbound}
Let $P$ and $Q$ be two mass functions on $\Xi \times \xspace$ such that $P(x | \xi) = Q(x | \xi)$ for all outcomes $\tuple{\xi,x}$. Then for all $x$ with $P(x)>0$:
\begin{equation}\label{eq:transferbound}
 -\log \frac{Q(x)}{P(x)} 
~\le~  E_P \sbr{-\log \frac{ Q(\rv \xi)}{P(\rv \xi)} \bigg| x}
~\le~ \max_\xi - \log \frac{Q(\xi)}{P(\xi)}. 
\end{equation}
Observe that if $Q(x) = 0$, we have $\infty \le  \infty \le \infty$.
\end{lemma}

\begin{proof}
For non-negative $a_1, \ldots a_m$ and $b_1, \ldots b_m$:
\begin{equation}\label{eq:logsum}
\del{\sum_{i=1}^m a_i} \log \frac{\sum_{i=1}^m a_i}{\sum_{i=1}^m b_i} 
~\le~ 
\sum_{i=1}^m a_i \log \frac{a_i}{b_i} 
~\le~ 
\del{\sum_{i=1}^m a_i} \max_i \log \frac{a_i}{b_i}.
\end{equation}
The first inequality is the log sum inequality~\cite[Theorem 2.7.1]{cover1991}. The second inequality is a simple overestimation. We now apply~\eqref{eq:logsum} substituting $m \mapsto \card{\Xi}$, $a_{\xi} \mapsto P(x,\xi)$ and $b_{\xi} \mapsto Q(x,\xi)$ and divide by $\sum_{i=1}^m a_i$ to obtain
\begin{equation*}
 -\log \frac{Q(x)}{P(x)}
~\le~
- \sum_{\xi} P(\xi|x) \log \frac{Q(\xi)}{P(\xi)}
~\le~ 
\max_{\xi} - \log \frac{Q(\xi)}{P(\xi)}. \qedhere
\end{equation*}
\end{proof}

\begin{proof}[Proof of Theorem~\ref{thm:unimix}]
  We first use \lemref{lemma:transferbound} to obtain a bound that
  does not depend on the data. Applying the lemma to the joint space
  $\xspace^n \times \Xi^n$, with
\begin{align*}
P(x^n,\xi^n) &\mapsto P_{\xi^n}(x^n)\pi_{\text{mix},\hat\alpha}(\xi^n) \quad \text{and} \\
Q(x^n,\xi^n) &\mapsto P_{\xi^n}(x^n)\pi_\text{umix}(\xi^n),
\end{align*}
yields loss bound
\begin{equation}\label{eq:unibnd1}
- \log P_\textnormal{umix}(x^n) - \hat L ~\le~ \max_{\xi^n} \del{-\log
  \pi_\textnormal{umix}(\xi^n) + \log
  \pi_{\textnormal{mix},\hat\alpha}(\xi^n)}.
\end{equation}
This bound can be computed prior to observation and without reference
to the experts' PFSs. The next step is to approximate the loss of
$\pi_\text{umix}$, which is itself well-known to be universal for the
multinomial distributions. It is shown in e.g. \cite{xiebarron2000}
that
\[-\log\pi_\text{umix}(\xi^n) \le -\log\pi_{\text{mix},\hat\alpha}(\xi^n)+{\card{\Xi}-1\over2}\log{n\over\pi}+c\]
for a fixed constant $c$. Combination with \eqref{eq:unibnd1} completes
the proof.
\end{proof}
\noindent Since the overhead incurred as a penalty for not knowing the
optimal parameter $\hat\alpha$ in advance is only logarithmic, we find
that $P_\text{umix}$ is strongly universal for the fixed elementwise
mixtures.

\subsubsection{HMM}
\begin{figure}
\centering
$\xymatrix@R=0.7em{
&&&&&&&&\\
&&&&&&&\ex{a} \ar@{.>}[ur]&\\
&&&&&&\si \name{0,3} \ar[ur] \ar[dr]&&\\
&&&&&\ex{a} \ar[ur]&&\ex{b} \ar@{.>}[dr]&\\
&&&&\si \name{0,2} \ar[ur] \ar[dr]&&&&\\
&&&\ex{a} \ar[ur]&&\ex{b} \ar[dr]&&\ex{a} \ar@{.>}[ur]&\\
&&\si \name{0,1} \ar[ur] \ar[dr]&&&&\si \name{\expname d,2} \ar[ur] \ar[dr]&&\\
&\ex{a} \ar[ur]&&\ex{b} \ar[dr]&&\ex{a} \ar[ur]&&\ex{b} \ar@{.>}[dr]&\\
\si \name{0,0} \ar[ur] \ar[dr]&&&&\si \name{\expname d,1} \ar[ur] \ar[dr]&&&&\\
&\ex{b} \ar[dr]&&\ex{a} \ar[ur]&&\ex{b} \ar[dr]&&\ex{a} \ar@{.>}[ur]&\\
&&\si \name{\expname d,0} \ar[ur] \ar[dr]&&&&\si \name{\expname c,1} \ar[ur] \ar[dr]&&\\
&&&\ex{b} \ar[dr]&&\ex{a} \ar[ur]&&\ex{b} \ar@{.>}[dr]&\\
&&&&\si \name{\expname c,0} \ar[ur] \ar[dr]&&&&\\
&&&&&\ex{b} \ar[dr]&&\ex{a} \ar@{.>}[ur]&\\
&&&&&&\si \name{\expname b,0} \ar[ur] \ar[dr]&&\\
&&&&&&&\ex{b} \ar@{.>}[dr]&\\
&&&&&&&&
}$
\caption{Combination of two experts using a universal elementwise mixture}\label{graph:unimix}
\end{figure}
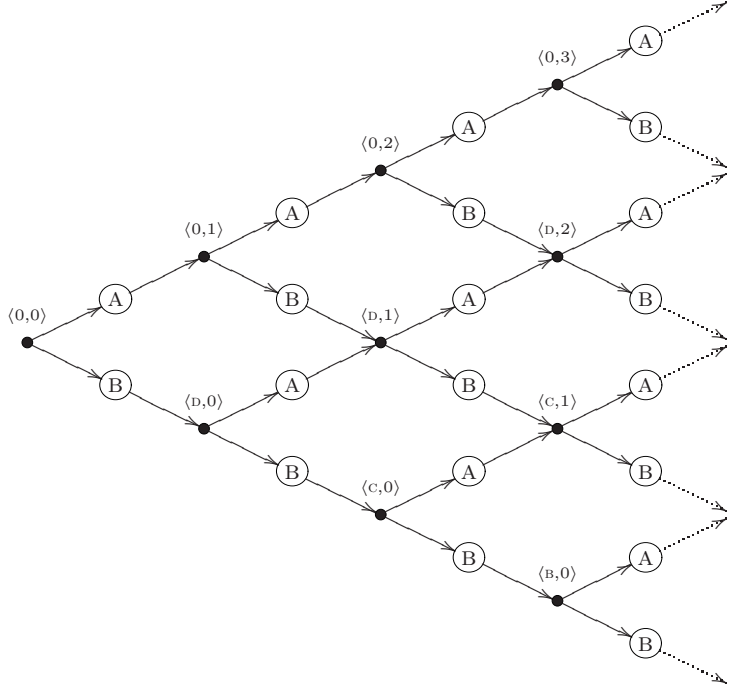

While universal elementwise mixtures can be described using the
ES-prior $\pi_\text{umix}$ defined in~\eqref{eq:umix}, unfortunately
any HMM that computes it needs a state for each possible count vector,
and is therefore huge if the number of experts is large. The HMM $\A_\text{umix}$ for
an arbitrary number of experts using the $\del{\tfrac{1}{2}, \ldots,
  \tfrac{1}{2}}$-Dirichlet prior is given using $\wstates = \wstates_\sil \cup \wstates_\prd$ by
\begin{gather}
\begin{aligned}
\wstates_\sil &= \nats^\Xi
&
\wstates_\prd &= \nats^\Xi \times \Xi
&
\winit(\mathbf 0) &= 1 
&
\wnl(\vec n, \xi) &= \xi 
\end{aligned}
\\
\label{eq:umix.wtf}
\wtf\del{
\begin{aligned}
\tuple{\vec n} &\to \tuple{\vec n, \xi}\\
\tuple{\vec n, \xi} &\to \tuple{\vec n + \mathbf 1_\xi}
\end{aligned}
} = \del{
\begin{gathered}
\tfrac{\wfrac{1}{2} + n_\xi}{\card{\Xi}/2 + \sum_\xi n_\xi} \\
1\\
\end{gathered}}
\end{gather}
We write $\nats^\Xi$ for the set of assignments of counts to experts; $\mathbf 0$ for the all zero assignment, and $\mathbf 1_\xi$ marks one count for expert $\xi$. We show the diagram of $\A_\text{umix}$ for the practical limit of two experts in \figref{graph:unimix}. In this case, the forward algorithm has running time $O(n^2)$.
Each productive state in \figref{graph:unimix} corresponds to a vector of two counts $(n_1, n_2)$ that sum to the sample size $n$, with the interpretation that of the $n$ experts, the first was used $n_1$ times while the second was used $n_2$ times. These counts are a sufficient statistic for the multinomial model class: per~\eqref{eq:pi.mix}~and~\eqref{eq:umix} the probability of the next expert only depends on the counts, and these probabilities are exactly the successor probabilities of the silent states \eqref{eq:umix.wtf}.

Other priors on $\alpha$ are possible. In particular, when all mass is placed on a single value of $\alpha$, we retrieve the elementwise mixture with fixed coefficients.

\subsection{Fixed Share}\label{sec:fixedshare}
\begin{figure}
\centering
$\xymatrix@R=0.7em@C=2.5em{
&\ex{a}\ar[dddr] \ar[rr]&&\ex{a}\ar[dddr] \ar[rr]&&\ex{a}\ar[dddr] \ar[rr]&&\ex{a}\ar@{.>}[r] \ar@{.>}[dddr]&\\
&&&&&&&\\
&\ex{b}\ar[dr]\ar[rr]&&\ex{b}\ar[dr]\ar[rr]&&\ex{b}\ar[dr]\ar[rr]&&\ex{b}\ar@{.>}[r] \ar@{.>}[dr]&\\
\si\name[+<0pt,11pt>]{\textsf{p},0} \ar[uuur]\ar[ur]\ar[dr]\ar[dddr]&&
\si\name[+<0pt,11pt>]{\textsf{p},1} \ar[uuur]\ar[ur]\ar[dr]\ar[dddr]&&
\si\name[+<0pt,11pt>]{\textsf{p},2} \ar[uuur]\ar[ur]\ar[dr]\ar[dddr]&&
\si\name[+<0pt,11pt>]{\textsf{p},3} \ar[uuur]\ar[ur]\ar[dr]\ar[dddr]&&
\\
&\ex{c}\ar[ur]\ar[rr]&&\ex{c}\ar[ur]\ar[rr]&&\ex{c}\ar[ur]\ar[rr]&&\ex{c}\ar@{.>}[r] \ar@{.>}[ur]&\\
&&&&&&&\\
&\ex{d}\ar[uuur]\ar[rr]&&\ex{d}\ar[uuur]\ar[rr]&&\ex{d}\ar[uuur]\ar[rr]&&\ex{d}\ar@{.>}[r] \ar@{.>}[uuur]&
}$
\caption{Combination of four experts using the fixed share algorithm}\label{graph:mixedshare}
\end{figure}
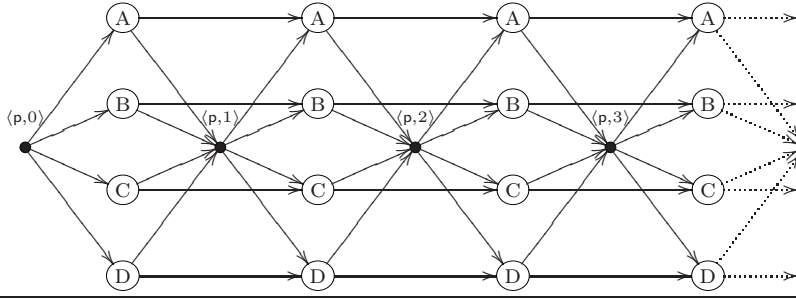

The first publication that considers a scenario where the best
predicting expert may change with the sample size is Herbster and
Warmuth's paper on \emph{tracking the best expert}
\cite{HerbsterWarmuth1995, HerbsterWarmuth1998}. They partition the
data of size $n$ into $m$ segments, where each segment is associated
with an expert, and give algorithms to predict almost as well as the
best \emph{partition} where the best expert is selected per segment.
They give two algorithms called fixed share and dynamic share. The
second algorithm does not fit in our framework; furthermore its motivation
applies only to loss functions other than log-loss. We focus on
fixed share, which is in fact identical to our algorithm applied to the
HMM depicted in \figref{graph:mixedshare}, where all arcs \emph{into}
the silent states have fixed probability $\alpha\in[0,1]$ and all arcs
\emph{from} the silent states have some fixed distribution $w$ on
$\Xi$.\footnote{This is actually a slight generalisation: the original
  algorithm uses a uniform $w(\xi)=1/\card{\Xi}$.} The same algorithm is
also described as an instance of the Aggregating Algorithm
in~\cite{Vovk1999}. Fixed share reduces to fixed elementwise
mixtures by setting $\alpha=1$ and to Bayesian mixtures by setting
$\alpha=0$. Formally:
\begin{subequations}
\begin{gather}
\begin{aligned}
\wstates &= \Xi \times \posints \cup \set{\textsf{p}} \times \nats
&
\winit(\textsf{p},0) &= 1
\\
\wstates_\prd &= \Xi \times \posints
&
\wnl(\xi,n) &= \xi
\end{aligned}
\\
\wtf\del{\begin{aligned}
\tuple{\textsf{p}, n} &\to \tuple{\xi,n+1}
\\
\tuple{\xi, n} &\to \tuple{\textsf{p},n}
\\
\tuple{\xi, n} &\to \tuple{\xi,n+1}
\end{aligned}} 
= 
\del{\begin{gathered}
w(\xi) 
\\
\alpha
\\
1-\alpha
\end{gathered}}
\end{gather}
\end{subequations}
Each productive state represents that a particular expert is used at a
certain sample size. Once a transition to a silent state is made, all history is forgotten and a new expert is chosen
according to $w$.\footnote{\label{fn:reflex}Contrary to the original fixed share, we
  allow switching to the same expert. In the HMM framework this is
  necessary to achieve running-time $ O(n\card{\Xi})$. Under uniform
  $w$, non-reflexive switching with fixed rate $\alpha$ can be
  simulated by reflexive switching with fixed rate $\beta =
  \frac{\alpha\card{\Xi}}{\card{\Xi}-1}$ (provided $\beta \le 1$). For non-uniform $w$, the
  rate becomes expert-dependent.}

Let $\hat L$ denote the loss achieved by the best partition, with
switching rate $\alpha^*:=m/(n-1)$.
Let $L_{\text{fs},\alpha}$ denote the loss of fixed share with uniform
$w$ and parameter $\alpha$. Herbster and Warmuth prove\footnote{
This bound can be obtained for the fixed share HMM using the previous footnote.
}
\[
L_{\text{fs},\alpha}-\hat L
~\le~
(n-1) H(\alpha^*, \alpha)+(m-1)\log(\card{\Xi}-1) + \log \card{\Xi},
\]
which we for brevity loosen slightly to
\begin{equation}\label{eq:fixedshare.lossbound}
L_{\text{fs},\alpha}-\hat L
~\le~
n H(\alpha^*, \alpha) +m\log \card{\Xi}.
\end{equation}
Here $H(\alpha^*,\alpha)=-\alpha^*\log\alpha-(1-\alpha^*)\log(1-\alpha)$ is the
cross entropy. The best loss guarantee is obtained for
$\alpha=\alpha^*$, in which case the cross entropy reduces to the binary entropy $H(\alpha)$.
A drawback of the method is that the optimal value of $\alpha$ has to
be known in advance in order to minimise the loss.  In
Sections~\secref{sec:universal.share} and \secref{sec:newstuff} we
describe a number of generalisations of fixed share that avoid this
problem.

\subsection{Universal Share}\label{sec:universal.share}
Independently, Volf and Willems describe universal share (they call it \emph{the switching method}) \cite{volfwillems1998}, which is very similar to a probabilistic
version of Herbster and Warmuth's fixed share algorithm,
except that they put a prior on the unknown parameter, with the result
that their algorithm adaptively learns the optimal value during
prediction. 

In \cite{bousquet2003}, Bousquet shows that the overhead for not knowing the optimal parameter value is equal to the overhead of a Bernoulli universal distribution. Let $L_{\text{fs},\alpha} = - \log P_{\text{fs},\alpha}(x^n)$ denote the loss achieved by the fixed share algorithm with parameter $\alpha$ on data $x^n$, and let $L_\text{us} = - \log P_\text{us}(x^n)$ denote the loss of universal share, where $P_\text{us}(x^n) = \int P_{\text{fs},\alpha}(x^n) w(\alpha) \dif \alpha$ with Jeffreys' prior $w(\alpha) = \alpha^{-1/2}(1-\alpha)^{-1/2}/\pi$ on $\intcc{0}{1}$. Then
\begin{equation}\label{eq:sm.loss.bound}
L_\text{us} - \min_\alpha L_{\text{fs},\alpha} \le 1+ \tfrac{1}{2}\log n.
\end{equation}
Thus $P_\text{us}$ is universal for the model class $\set{P_{\text{fs},\alpha} \mid \alpha \in \intcc{0}{1}}$ that consists of all ES-joints where the ES-priors are distributions with a fixed switching rate.

Universal share requires quadratic running time $O(n^2\card{\Xi})$, restricting its use to moderately small data sets.

In~\cite{Jaakkola2003}, Monteleoni
and Jaakkola place a discrete prior on the parameter that divides its mass over $\sqrt n$ well-chosen points, in a setting where the ultimate sample size $n$ is known beforehand. This way they still manage to achieve \eqref{eq:sm.loss.bound} up to a constant, while reducing the running time to $O(n\sqrt n \card{\Xi})$.

In~\cite{bousquet2002}, Bousquet and Warmuth describe
yet another generalisation of expert tracking; they derive good loss
bounds in the situation where the best experts for each section in the
partition are drawn from a small pool.

\begin{figure}
\centering
$\xymatrix@R=0.7em{
&&&&&&&&&&\\
\\
\\
\\
\\
\\
&&&&\ex{a} \name{\expname a,1,2} \ar[dddr] \ar[rrr]&&&\ex{a}\ar[dddr] \ar[rrr]&&&\ex{a}\ar@{.>}[r] \ar@{.>}[dddr]&\\
&&&&&&&\\
&&&&\ex{b} \name{\expname b,1,2} \ar[dr]\ar[rrr]&&&\ex{b}\ar[dr]\ar[rrr]&&&\ex{b}\ar@{.>}[r] \ar@{.>}[dr]&\\
&&&
\si\name[+<-7pt,10pt>]{\textsf{p},1,1}\ar[uuur]\ar[ur]\ar[dr]\ar[dddr]&&
\si\name[+<6pt,-11pt>]{\textsf{q},1,2}\ar@{.>}[uuuuuuuuur]&\si\name[+<-7pt,10pt>]{\textsf{p},1,2}\ar[uuur]\ar[ur]\ar[dr]\ar[dddr]&&
\si\name[+<6pt,-11pt>]{\textsf{q},1,3}\ar@{.>}[uuuuuuuuur]&\si\name[+<-7pt,10pt>]{\textsf{p},1,3}\ar[uuur]\ar[ur]\ar[dr]\ar[dddr]&&
\\
&&&&\ex{c} \name{\expname c,1,2} \ar[ur]\ar[rrr]&&&\ex{c}\ar[ur]\ar[rrr]&&&\ex{c}\ar@{.>}[r] \ar@{.>}[ur]&\\
&&&&&&&\\
&&&&\ex{d} \name[+<0pt,15pt>]{\expname d,1,2} \ar[uuur]\ar[rrr]&&&\ex{d}\ar[uuur]\ar[rrr]&&&\ex{d}\ar@{.>}[r] \ar@{.>}[uuur]&
\\
\\
\\
&\ex{a} \name{\expname a,0,1}\ar[dddr] \ar[rrr]&&&\ex{a}\ar[dddr] \ar[rrr]&&&\ex{a}\ar[dddr] \ar[rrr]&&&\ex{a}\ar@{.>}[r] \ar@{.>}[dddr]&\\
&&&&&&&\\
&\ex{b} \name{\expname b,0,1} \ar[dr]\ar[rrr]&&&\ex{b}\ar[dr]\ar[rrr]&&&\ex{b}\ar[dr]\ar[rrr]&&&\ex{b}\ar@{.>}[r] \ar@{.>}[dr]&\\
\si\name[+<-7pt,10pt>]{\textsf{p},0,0}\ar[uuur]\ar[ur]\ar[dr]\ar[dddr]&&
\si\name[+<6pt,-11pt>]{\textsf{q},0,1}\ar[uuuuuuuuur]&&&
\si\name[+<6pt,-11pt>]{\textsf{q},0,2}\ar[uuuuuuuuur]&&&
\si\name[+<6pt,-11pt>]{\textsf{q},0,3}\ar[uuuuuuuuur]&&&
\\
&\ex{c}\name{\expname c,0,1} \ar[ur]\ar[rrr]&&&\ex{c}\ar[ur]\ar[rrr]&&&\ex{c}\ar[ur]\ar[rrr]&&&\ex{c}\ar@{.>}[r] \ar@{.>}[ur]&\\
&&&&&&&\\
&\ex{d}\name[+<0pt,15pt>]{\expname d,0,1}\ar[uuur]\ar[rrr]&&&\ex{d}\ar[uuur]\ar[rrr]&&&\ex{d}\ar[uuur]\ar[rrr]&&&\ex{d}\ar@{.>}[r] \ar@{.>}[uuur]&
}$
\caption{Combination of four experts using universal share}\label{graph:universal.share}
\end{figure}
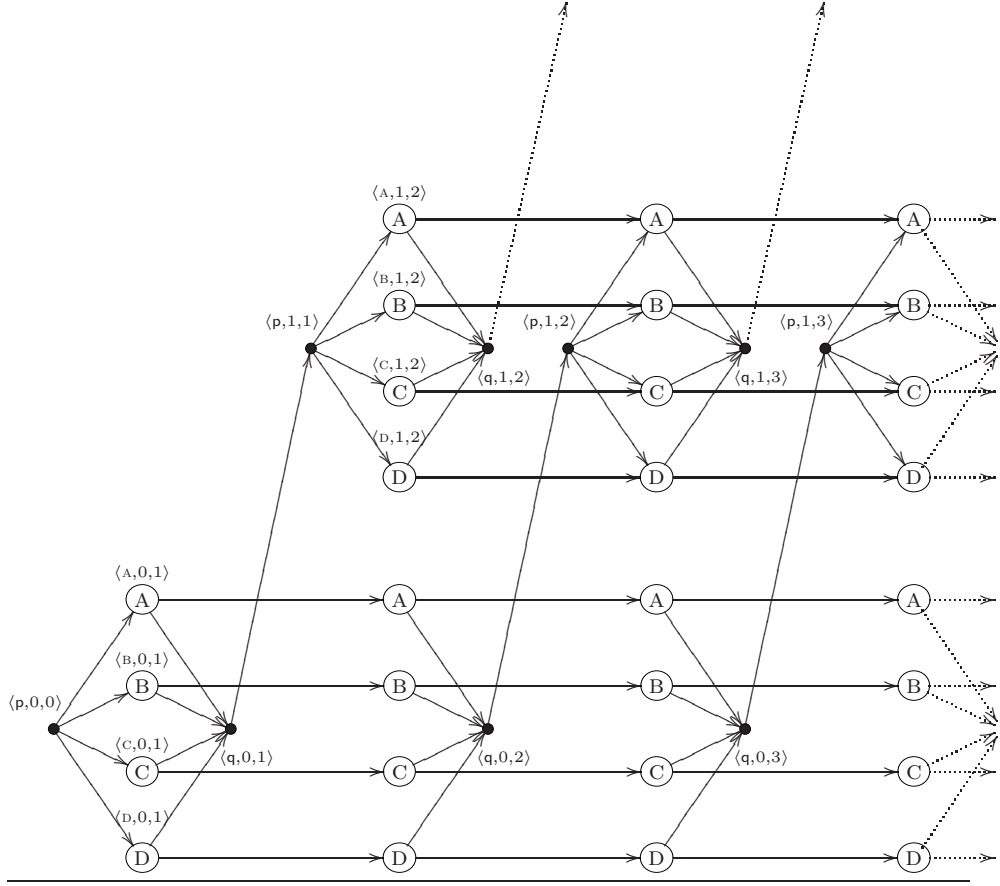

The HMM for universal share with the $\del{\tfrac{1}{2}, \tfrac{1}{2}}$-Dirichlet prior on the switching rate $\alpha$ is displayed in \figref{graph:universal.share}.  It is formally specified (using $\wstates = \wstates_\sil \cup \wstates_\prd$) by:
\begin{subequations}
\begin{gather}
\begin{array}{r@{~}c@{~}l}
\wstates_\sil = &\set{\textsf{p},\textsf{q}} &\times \set{\tuple{m,n} \in \nats^2 \mid m \le n} \\
\wstates_\prd = &\Xi &\times \set{\tuple{m,n} \in \nats^2 \mid m < n}
\end{array}
\\
\begin{aligned}
\wnl(\xi,m,n) &= \xi
&\qquad
\winit(\textsf{p},0,0) &= 1
\end{aligned}
\\
\label{eq:sw.method.wtf.def}
\wtf \del{
\begin{aligned}
\tuple{\textsf{p},m,n} &\to \tuple{\xi,m,n+1}
\\
\tuple{\textsf{q},m,n} &\to \tuple{\textsf{p},m+1,n}
\\
\tuple{\xi, m,n} &\to \tuple{\textsf{q},m,n}
\\
\tuple{\xi, m,n} &\to \tuple{\xi,m,n+1}
\end{aligned}} =
\del{
\begin{gathered}
w(\xi) 
\\
1
\\
\wfrac{(m+\tfrac{1}{2})}{n}
\\
\wfrac{(n-m-\tfrac{1}{2})}{n}
\end{gathered}}
\end{gather}
\end{subequations}
\noindent Each productive state $\tuple{\xi, n, m}$ represents the fact that at sample size $n$ expert $\xi$ is used, while there have been $m$ switches in the past. Note that the last two lines of \eqref{eq:sw.method.wtf.def} are subtly different from the corresponding topmost line of \eqref{eq:umix.wtf}. In a sample of size $n$ there are $n$ possible positions to use a given expert, while there are only $n-1$ possible switch positions. 

The presence of the switch count in the state is the new ingredient compared to fixed share. It allows us to adapt the switching probability to the data, but it also renders the number of states quadratic. We discuss reducing the number of states without sacrificing much performance in \secref{sec:alg.tricks}.

\subsection{Overconfident Experts}\label{sec:overconfident}
In~\cite{Vovk1999}, Vovk considers overconfident experts. In this
scenario, there is a single unknown best expert, except that this
expert sometimes makes wild (over-categorical) predictions. We assume
that the rate at which this happens is a known constant $\alpha$. The
overconfident expert model is an attempt to mitigate the wild
predictions using an additional ``safe'' expert $\expname u \in \Xi$,
who always issues the uniform distribution on $\xspace$ (which we
assume to be finite for simplicity here). Using $\wstates =
\wstates_\sil \cup \wstates_\prd$, it is formally specified by:
\begin{subequations}
\begin{gather}
\begin{aligned}
\wstates_\sil &= \Xi \times \nats
&
\wnl(\textsf n, \xi, n) &= \xi
&
\winit(\xi,0) &= w(\xi)
\\
\wstates_\prd &= \set{\textsf n,\textsf w} \times \Xi \times \posints
&
\wnl(\textsf w, \xi, n) &= u 
\end{aligned}
\\
\wtf\del{\begin{aligned}
\tuple{\xi,n} &\to \tuple{\textsf n, \xi, n+1}
\\
\tuple{\xi,n} &\to \tuple{\textsf w, \xi, n+1}
\\
\tuple{\textsf n, \xi, n} &\to \tuple{\xi,n}
\\
\tuple{\textsf w, \xi, n} &\to \tuple{\xi,n}
\end{aligned}}
= \del{
\begin{gathered}
1-\alpha
\\
\alpha
\\
1
\\
1
\end{gathered}}
\end{gather}
\end{subequations}
\begin{figure}
\centering
$\xymatrix@R=0.7em{
&&\ex{a}\ar[dr]&&\ex{a}\ar[dr]&&\ex{a}\ar[dr]&&\ex{a}\ar@{.>}[dr]&\\
&\si \name{\expname a,0} \ar[ur]\ar[dr]&&\si \name{\expname a,1} \ar[ur]\ar[dr]&&\si \ar[ur]\ar[dr]&&\si \ar[ur]\ar[dr]&&\\
&&\ex{u}\ar[ur]&&\ex{u}\ar[ur]&&\ex{u}\ar[ur]&&\ex{u}\ar@{.>}[ur]&\\
&&\\
&&\ex{b}\name{\textsf{n},3,1} \ar[dr]&&\ex{b}\ar[dr]&&\ex{b}\ar[dr]&&\ex{b}\ar@{.>}[dr]&\\
&\si \name{\expname b,0} \ar[ur]\ar[dr]&&\si \name{\expname b,1} \ar[ur]\ar[dr]&&\si \ar[ur]\ar[dr]&&\si \ar[ur]\ar[dr]&&\\
&&\ex{u}\name{\textsf{w},3,1} \ar[ur]&&\ex{u}\ar[ur]&&\ex{u}\ar[ur]&&\ex{u}\ar@{.>}[ur]&\\
\bn \ar[uur] \ar[ddr] \ar[uuuuuur] \ar[ddddddr]&&\\
&&\ex{c}\ar[dr]&&\ex{c}\ar[dr]&&\ex{c}\ar[dr]&&\ex{c}\ar@{.>}[dr]&\\
&\si \name[+<2pt,10pt>]{\expname c,0} \ar[ur]\ar[dr]&&\si \name{\expname c,1} \ar[ur]\ar[dr]&&\si \ar[ur]\ar[dr]&&\si \ar[ur]\ar[dr]&&\\
&&\ex{u}\ar[ur]&&\ex{u}\ar[ur]&&\ex{u}\ar[ur]&&\ex{u}\ar@{.>}[ur]&\\
&&\\
&&\ex{d}\ar[dr]&&\ex{d}\ar[dr]&&\ex{d}\ar[dr]&&\ex{d}\ar@{.>}[dr]&\\
&\si \name[+<5pt,12pt>]{\expname d,0} \ar[ur]\ar[dr]&&\si \name{\expname d,1} \ar[ur]\ar[dr]&&\si \ar[ur]\ar[dr]&&\si \ar[ur]\ar[dr]&&\\
&&\ex{u}\ar[ur]&&\ex{u}\ar[ur]&&\ex{u}\ar[ur]&&\ex{u}\ar@{.>}[ur]&\\
}$
\caption{Combination of four overconfident experts}\label{graph:overconfidentexperts}
\end{figure}
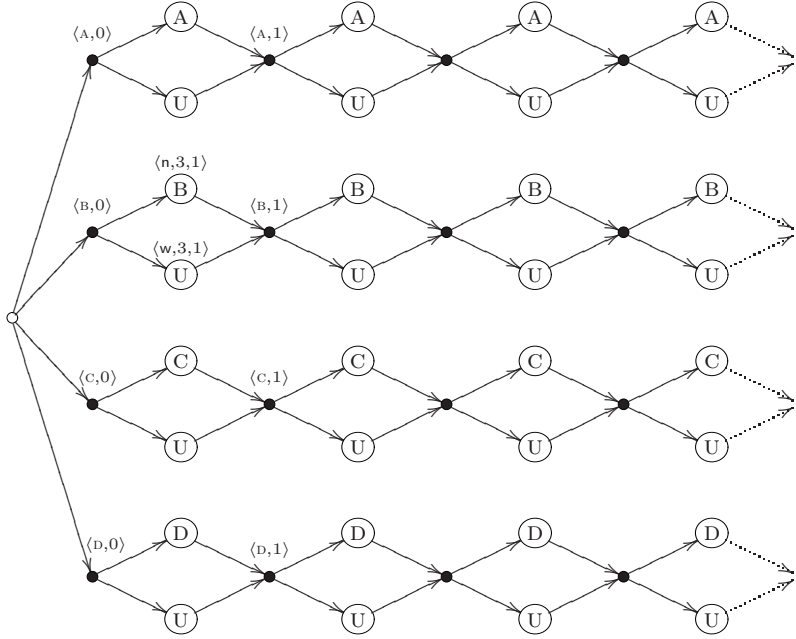%
Each productive state corresponds to the idea that a certain expert is
best, and additionally whether the current outcome is normal or wild.

Fix data $x^n$. Let $\hat\xi^n$ be the expert sequence that maximises the likelihood $P_{\xi^n}(x^n)$ among all expert sequences $\xi^n$ that switch between a single expert and $\expname u$. To derive our loss bound, we underestimate the marginal probability $P_{\text{oce},\alpha}(x^n)$ for the HMM defined above, by dropping all terms except the one for $\hat\xi^n$.
\begin{equation}
P_{\text{oce},\alpha}(x^n)\quad=~\sum_{\xi^n\in\Xi^n}\pi_{\text{oce},\alpha}(\xi^n)P_{\xi^n}(x^n)\quad\ge\quad\pi_{\text{oce},\alpha}(\hat\xi^n)P_{\hat\xi^n}(x^n).
\end{equation}
(This first step is also used in the bounds for the two new models in
\secref{sec:newstuff}.) 
Let $\alpha^*$ denote the frequency of occurrence
of $\expname u$ in $\hat\xi^n$, let $\xi_\text{best}$ be the other expert that occurs in $\xi^n$, and let $\hat L=-\log P_{\hat\xi^n}(x^n)$. We can now bound our worst-case additional loss:
\[
- \log P_{\text{oce},\hat\alpha}(x^n) - \hat L \le - \log
\pi_{\text{oce},\alpha}(\hat\xi^n)=-\log w(\xi_\text{best})+nH(\alpha^*,\alpha).
\]
Again $H$ denotes the cross entropy. From a coding perspective, after first
specifying the best expert $\xi_\text{best}$ and a binary sequence
representing $\hat\xi^n$, we can then use $\hat\xi^n$ to encode the
actual observations with optimal efficiency.

The optimal misprediction rate $\alpha$ is usually not known in
advance, so we can again learn it from data by placing a prior on it
and integrating over this prior. This comes at the cost of an
additional loss of ${1\over2}\log n+c$ bits for some constant $c$
(which is $\le 1$ for two experts), and as will be shown in the next
subsection, can be implemented using a quadratic time algorithm.

\subsubsection{Recursive Combination}\label{sec:rec.comb}
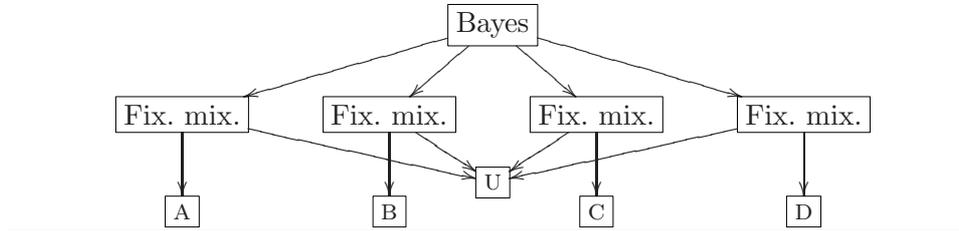
\begin{figure}
\caption{Implementing overconfident experts with recursive
  combinations.}\label{fig:overconfident.rec.comb.app}
\centerline{
\xymatrix@C=1em{
&&&  \save *+[F]\txt{Bayes} \ar[dlll] \ar[dl] \ar[dr] \ar[drrr]  \restore
\\
 *+[F]\txt{Fix.\ mix.}  \ar[d] 
&
& *+[F]\txt{Fix.\ mix.} \ar[d]
&
& *+[F]\txt{Fix.\ mix.} \ar[d]
&
& *+[F]\txt{Fix.\ mix.} \ar[d]
\\
  *+[F]{\expname a} 
&
& *+[F]{\expname b} 
& \save []+<0cm,1em> *+[F]{\expname u} \ar@{<-}[ulll] \ar@{<-}[ul] \ar@{<-}[ur] \ar@{<-}[urrr] \restore
& *+[F]{\expname c}
& 
& *+[F]{\expname d}
}}
\end{figure}

In \figref{graph:overconfidentexperts} one may recognise two simpler
HMMs: it is in fact just a Bayesian combination of a set of fixed
elementwise mixtures with some parameter $\alpha$, one for each
expert. Thus two models for combining expert predictions, the Bayesian
model and fixed elementwise mixtures, have been recursively combined
into a single new model. This view is illustrated in
\figref{fig:overconfident.rec.comb.app}.

More generally, any method to combine the predictions of multiple
experts into a single new prediction strategy, can itself be
considered an expert. We can apply our method recursively to this new
``meta-expert''; the running time of the recursive combination is only
the \emph{sum} of the running times of all the component predictors.
For example, if all used individual expert models can be evaluated in
quadratic time, then the full recursive combination also has quadratic
running time, \emph{even though it may be impossible to specify using
  an HMM of quadratic size}.

Although a recursive combination to implement overconfident experts
may save some work, the same running time may be achieved by
implementing the HMM depicted in \figref{graph:overconfidentexperts}
directly. However, we can also obtain efficient generalisations of the
overconfident expert model, by replacing any combinator by a more
sophisticated one. For example, rather than a fixed elementwise
mixture, we could use a universal elementwise mixture for each expert,
so that the error frequency is learned from data. Or, if we suspect
that an expert may not only make incidental slip-ups, but actually
become completely untrustworthy for longer stretches of time, we may
even use a fixed or universal share model.

One may also consider that the fundamental idea behind the
overconfident expert model is to combine each expert with a uniform
predictor using a misprediction model. In the example in
\figref{fig:overconfident.rec.comb.app}, this idea is used to
``smooth'' the expert predictions, which are then used at the top
level in a Bayesian combination. However, the model that is used at
the top level is completely orthogonal to the model used to smooth
expert predictions; we can safeguard against overconfident experts not
only in Bayesian combinations but also in other models such as the
switch distribution or the run-length model, which are described in
the next section.

\section{New Models to Switch between Experts}\label{sec:newstuff}
So far we have considered two models for switching between experts: fixed share and its generalisation, universal share. While fixed share is an extremely efficient algorithm, it requires that the frequency of switching between experts is estimated a priori, which can be hard in practice. Moreover, we may have prior knowledge about how the switching probability will change over time, but unless we know the ultimate sample size in advance, we may be forced to accept a linear overhead compared to the best parameter value. Universal share overcomes this problem by marginalising over the unknown parameter, but has quadratic running time.

The first model considered in this section, called the switch distribution, avoids both problems. It is parameterless and has essentially the same running time as fixed share. It also achieves a loss bound competitive to that of universal share. Moreover, for a bounded number of switches the bound has even better asymptotics.

The second model is called the run-length model because it uses a run-length code (c.f.\ \cite{Moffat2002}) as an ES-prior. This may be useful because, while both fixed and universal share model the distance between switches with a geometric distribution, the real distribution on these distances may be different. This is the case if, for example, the switches are highly clustered. This additional expressive power comes at the cost of quadratic running time, but we discuss a special case where this may be reduced to linear. 

We conclude this section with a comparison of the four expert switching models discussed in this paper.

\subsection{Switch Distribution}\label{sec:switch}
The switch distribution is a new model for combining expert
predictions. Like fixed share, it is intended for settings where the
best predicting expert is expected to change as a function of the
sample size, but it has two major innovations. First, we let the
probability of switching to a different expert decrease with the
sample size. This allows us to derive a loss bound close to that of
the fixed share algorithm, without the need to tune any
parameters.\footnote{The idea of decreasing the switch probability as $1/(n+1)$,
  which has not previously been published, was independently conceived
  by Mark Herbster and the authors.} Second, the switch distribution
has a special provision to ensure that in the case where the number of
switches remains bounded, the incurred loss overhead is $O(1)$.

The switch distribution was introduced in~\cite{threemusketeers07},
which addresses a long standing open problem in statistical model
class selection known as the ``AIC vs BIC dilemma''. Some criteria
for model class selection, such as AIC, are efficient when applied to
sequential prediction of future outcomes, while other criteria, such
as BIC, are ``consistent'': with probability one, the model class that
contains the data generating distribution is selected given enough
data. Using the switch distribution, these two goals (truth finding vs
prediction) can be reconciled. Refer to the paper for more
information.

Here we disregard such applications and treat the switch distribution
like the other models for combining expert predictions. We describe an
HMM that corresponds to the switch distribution; this illuminates the
relationship between the switch distribution and the fixed share
algorithm which it in fact generalises.

The equivalence between the original
definition of the switch distribution and the HMM is not trivial, so
we give a formal proof. The size of the HMM is such that calculation
of $P(x^n)$ requires only $O(n\card{\Xi})$ steps.

We provide a loss bound for the switch distribution in
\secref{sec:switchbound}. Then in \secref{sec:switch.map} we show how the
sequence of experts that has maximum a posteriori probability can be
computed. This problem is difficult for general HMMs, but the
structure of the HMM for the switch distribution allows for an
efficient algorithm in this case.

\subsubsection{Switch HMM}\label{sec:switch.hmm}
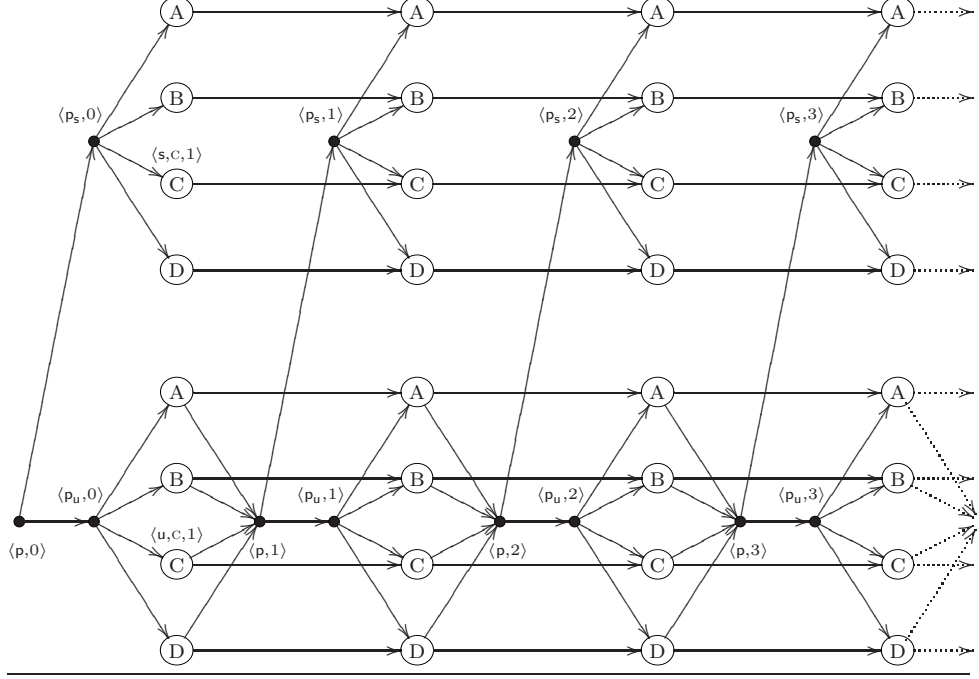
\begin{figure}
\centering
$\xymatrix@R=0.7em@C=2em{
&&\ex{a} \ar[rrr]&&&\ex{a} \ar[rrr]&&&\ex{a} \ar[rrr]&&&\ex{a}\ar@{.>}[r] &\\
\\
&&\ex{b}\ar[rrr]&&&\ex{b}\ar[rrr]&&&\ex{b}\ar[rrr]&&&\ex{b}\ar@{.>}[r] &\\
&\si \name[+<-5pt,10pt>]{$\textsf{p}_\textsf{s}$,0} \ar[uuur]\ar[ur]\ar[dr]\ar[dddr]&&
&\si \name[+<-5pt,10pt>]{$\textsf{p}_\textsf{s}$,1} \ar[uuur]\ar[ur]\ar[dr]\ar[dddr]&&
&\si \name[+<-5pt,10pt>]{$\textsf{p}_\textsf{s}$,2} \ar[uuur]\ar[ur]\ar[dr]\ar[dddr]&&
&\si \name[+<-5pt,10pt>]{$\textsf{p}_\textsf{s}$,3} \ar[uuur]\ar[ur]\ar[dr]\ar[dddr]&&\\
&&\ex{c}\name[+<0pt,11pt>]{\textsf{s},\expname c,1} \ar[rrr]&&&\ex{c}\ar[rrr]&&&\ex{c}\ar[rrr]&&&\ex{c}\ar@{.>}[r] &\\
\\
&&\ex{d} \ar[rrr]&&&\ex{d}\ar[rrr]&&&\ex{d}\ar[rrr]&&&\ex{d}\ar@{.>}[r] &
\\
\\
\\
&&\ex{a}  \ar[dddr] \ar[rrr]&&&\ex{a}\ar[dddr] \ar[rrr]&&&\ex{a}\ar[dddr] \ar[rrr]&&&\ex{a}\ar@{.>}[r] \ar@{.>}[dddr]&\\
\\
&&\ex{b}\ar[dr]\ar[rrr]&&&\ex{b}\ar[dr]\ar[rrr]&&&\ex{b}\ar[dr]\ar[rrr]&&&\ex{b}\ar@{.>}[r] \ar@{.>}[dr]&\\
\si \name[+<3pt,-11pt>]{\textsf{p},0} \ar[r] \ar[uuuuuuuuur]
&\si \name[+<-5pt,10pt>]{$\textsf{p}_\textsf{u}$,0}\ar[uuur]\ar[ur]\ar[dr]\ar[dddr]&&
\si \name[+<3pt,-11pt>]{\textsf{p},1} \ar[uuuuuuuuur]\ar[r]
&\si \name[+<-5pt,10pt>]{$\textsf{p}_\textsf{u}$,1}\ar[uuur]\ar[ur]\ar[dr]\ar[dddr]&&
\si \name[+<3pt,-11pt>]{\textsf{p},2} \ar[uuuuuuuuur]\ar[r]
&\si \name[+<-5pt,10pt>]{$\textsf{p}_\textsf{u}$,2}\ar[uuur]\ar[ur]\ar[dr]\ar[dddr]&&
\si \name[+<3pt,-11pt>]{\textsf{p},3} \ar[uuuuuuuuur]\ar[r]
&\si \name[+<-5pt,10pt>]{$\textsf{p}_\textsf{u}$,3}\ar[uuur]\ar[ur]\ar[dr]\ar[dddr]&&
\\
&&\ex{c} \name[+<0pt,11pt>]{\textsf{u},\expname c,1} \ar[ur]\ar[rrr]&&&\ex{c}\ar[ur]\ar[rrr]&&&\ex{c}\ar[ur]\ar[rrr]&&&\ex{c}\ar@{.>}[r] \ar@{.>}[ur]&\\
\\
&&\ex{d}\ar[uuur]\ar[rrr]&&&\ex{d}\ar[uuur]\ar[rrr]&&&\ex{d}\ar[uuur]\ar[rrr]&&&\ex{d}\ar@{.>}[r] \ar@{.>}[uuur]&
}$
\caption{Combination of four experts using the switch distribution}\label{graph:switch}
\end{figure}

Let $\sigma^\omega$ and $\tau^\omega$ be sequences of distributions on $\set{0,1}$ which we call the \markdef{switch probabilities} and the \markdef{stabilisation probabilities}. The switch HMM $\A_\sw$, displayed in \figref{graph:switch}, is defined below using $\wstates = \wstates_\sil \cup \wstates_\prd$:
\begin{subequations}\label{eq:switch.hmm}
\begin{gather}
\begin{aligned}
\wstates_\sil &= \set{\textsf{p}, \textsf{p}_{\textsf s}, \textsf{p}_{\textsf u}} \times \nats
&
\winit(\textsf{p}, 0) &= 1
&
\wnl(\textsf{s}, \xi, n) &= \xi
\\
\wstates_\prd &= \set{\textsf{s}, \textsf{u}} \times \Xi \times \posints
&
&&
\wnl(\textsf{u}, \xi, n) &= \xi
\end{aligned}
\\
\wtf\del{\begin{aligned}
\tuple{\textsf{p}, n} & \to  \tuple{\textsf{p}_{\textsf u}, n} 
\\
\tuple{\textsf{p}, n} & \to  \tuple{\textsf{p}_{\textsf s}, n}
\\
\tuple{\textsf{p}_{\textsf u}, n} & \to \tuple{\textsf{u}, \xi, n+1}
\\
\tuple{\textsf{p}_{\textsf s}, n} & \to \tuple{\textsf{s}, \xi, n+1}
\\
\tuple{\textsf{s}, \xi, n} & \to  \tuple{\textsf{s}, \xi, n+1} 
\\
\tuple{\textsf{u}, \xi, n} & \to \tuple{\textsf{u}, \xi, n+1}
\\
\tuple{\textsf{u}, \xi, n} & \to \tuple{\textsf{p}, n}
\end{aligned}} 
= 
\del{\begin{gathered}
\tau_n(0)
\\
\tau_n(1)
\\
w(\xi)
\\
w(\xi)
\\
1
\\
\sigma_n(0)
\\
\sigma_n(1)
\end{gathered}}
\end{gather}
\end{subequations}
This HMM contains two ``expert bands''. Consider a productive state $\tuple{\textsf{u}, \xi, n}$ in the bottom band, which we call the \emph{unstable} band, from a generative viewpoint. Two things can happen. With probability $\sigma_n(0)$ the process continues horizontally to $\tuple{\textsf{u}, \xi, n+1}$ and the story repeats. We say that \emph{no switch occurs}. With probability $\sigma_n(1)$ the process continues to the silent state $\tuple{\textsf{p},n}$ directly to the right. We say that \emph{a switch occurs}. Then a new choice has to be made. With probability $\tau_n(0)$ the process continues rightward to $\tuple{\textsf{p}_\textsf{u}, n}$ and then branches out to some productive state $\tuple{\textsf{u}, \xi', n+1}$ (possibly $\xi = \xi'$), and the story repeats. With probability $\tau_n(1)$ the process continues to $\tuple{\textsf{p}_\textsf{s}, n}$ in the top band, called the \emph{stable} band. Also here it branches out to some productive state $\tuple{\textsf{s}, \xi', n+1}$. But from this point onward there are no choices anymore; expert $\xi'$ is produced forever. We say that the process has \emph{stabilised}.

By choosing $\tau_n(1) = 0$ and $\sigma_n(1) = \theta$ for all $n$ we essentially remove the stable band and arrive at fixed share with parameter $\theta$. The presence of the stable band enables us to improve the loss bound of fixed share in the particular case that the number of switches is bounded; in that case, the stable band allows us to remove the dependency of the loss bound on $n$ altogether. We will use the particular choice $\tau_n(0) = \theta$ for all $n$, and $\sigma_n(1) = \pit(\rv Z=n|\rv Z \ge n)$ for some fixed value $\theta$ and an arbitrary distribution $\pit$ on $\nats$. This allows us to relate the switch HMM to the parametric representation that we present next.

\subsubsection{Switch Distribution}
In~\cite{threemusketeers07} De Rooij, Van Erven and Gr\"unwald
introduce a prior distribution on expert sequences and give an algorithm 
that computes it
efficiently, i.e.\ in time $O(n\card{\Xi})$, where $n$ is the sample
size and $\card{\Xi}$ is the number of considered experts. In this
section, we will prove that the switch distribution is implemented by
the switch HMM of \secref{sec:switch.hmm}. Thus, the algorithm given
in~\cite{threemusketeers07} is really just the forward algorithm
applied to the switch HMM.

\begin{definition} We first define the countable set of \markdef{switch parameters}
\[ \Theta_\sw := \set{\tuple{t^m, k^m} \mid m \ge 1, k \in \Xi^m, t \in \nats^m \text{ and } 0 = t_1 < t_2 < t_3 \ldots}.\] The \markdef{switch prior} is the discrete distribution on switch parameters given by
\[ \pi_\sw(t^m, k^m) := \pim(m) \pik(k_1) \prod_{i=2}^m \pit(t_i|t_i > t_{i-1}) \pik(k_i),\]
where $\pim$ is geometric with rate $\theta$, $\pit$ and $\pik$ are arbitrary distributions on $\nats$ and $\Xi$.
We define the mapping $\xi : \Theta_\sw \to \Xi^\omega$ that interprets switch parameters as sequences of experts by 
\[\xi(t^m, k^m) :=  k_1^{[t_2-t_1]} \conc k_2^{[t_3-t_2]} \conc \ldots \conc k_{m-1}^{[t_m-t_{m-1}]} \conc k_m^{[\omega]}, \]
where $k^{[\lambda]}$ is the sequence consisting of $\lambda$ repetitions of $k$. This mapping is not 1-1: infinitely many switch parameters map to the same infinite sequence, since $k_i$ and $k_{i+1}$ may coincide. The \markdef{switch distribution} $P_\sw$ is the ES-joint based on the ES-prior that is obtained by composing $\pi_\sw$ with $\xi$.
\end{definition}

\subsubsection{Equivalence}
In this section we show that the HMM prior $\pi_\A$ and the switch prior $\pi_\sw$ define the same ES-prior. During this section, it is convenient to regard $\pi_\A$ as a distribution on sequences of states, allowing us to differentiate between distinct sequences of states that map to the same sequence of experts. The function $\trace : \wstates^\omega \to \Xi^\omega$, that we call \markdef{trace}, explicitly performs this mapping; $\trace(\wstate^\omega)(i) := \wnl(\wstate^\prd_i)$. We cannot relate $\pi_\sw$ to $\pi_\A$ directly as they are carried by different sets (switch parameters vs state sequences), but need to consider the distribution that both induce on sequences of experts via $\xi$ and $\trace$. Formally:

\begin{definition}
If $f:\Theta\rightarrow\Gamma$ is a random variable and $P$ is a
distribution on $\Theta$, then we write $f(P)$ to denote the
distribution on $\Gamma$ that is induced by $f$.
\end{definition}

\begin{figure}
\[
\xymatrix{
\wstates^\omega,\pi_\A \ar[rdd]_{\trace} &
&\Theta_\sw,\pi_\sw \ar[dld]^\xi \ar[ll]_f \\
&\#&\\
 & \Xi^\omega,\pi & \\
}\]
\caption{Commutativity diagram}\label{fig:commutation_diagram}
\end{figure}
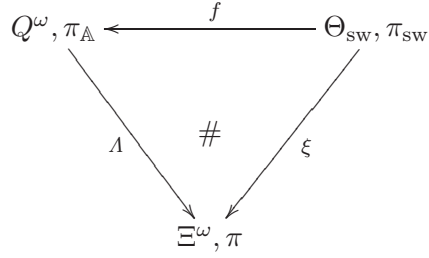

\noindent
Below we will show that $\trace(\pi_\A) = \xi(\pi_\sw)$, i.e.\ that $\pi_\sw$ and $\pi_\A$ induce the same distribution on the expert sequences $\Xi^\omega$ via the trace $\trace$ and the expert-sequence mapping $\xi$. Our argument will have the structure outlined in \figref{fig:commutation_diagram}. Instead of proving the claim directly, we create a random variable $f : \Theta_\sw \to \wstates^\omega$ mapping switch parameters into runs. Via $f$, we can view $\Theta_\sw$ as a reparametrisation of $\wstates^\omega$. We then show that the diagram commutes, that is, $\pi_\A = f(\pi_\sw)$ and $\trace \circ f = \xi$. This shows that $\trace(\pi_\A) = \trace(f(\pi_\sw)) = \xi(\pi_\sw)$ as required.

\begin{proposition} Let $\A$ be the HMM as defined in \secref{sec:switch.hmm}, and $\pi_\sw, \xi$ and $\trace$ as above. If $w = \pik$ then
\[  \xi(\pi_\sw) = \trace(\pi_\A).\]
\end{proposition}

\begin{proof} Recall \eqref{eq:switch.hmm} that 
\[\wstates = \set{\textsf{s}, \textsf{u}} \times \Xi \times \posints  \quad\cup\quad \set{\textsf{p}, \textsf{p}_{\textsf s}, \textsf{p}_{\textsf u}} \times \nats.
\] 
We define the random variable $f : \Theta_\sw \to \wstates^\omega$ by
\begin{align*}
f(t^m, k^m) &:=  \tuple{\textsf{p}, 0} \conc u_1 \conc u_2 \conc \ldots \conc u_{m-1} \conc s, \qquad\text{where}
\\
u_i &:= \tuple{\tuple{\textsf{p}_\textsf{u}, t_i}, \tuple{\textsf{u},k_i, t_i+1}, \tuple{\textsf{u},k_i,t_i+2}, \ldots, \tuple{\textsf{u},k_i,t_{i+1}}, \tuple{\textsf{p}, t_{i+1}}}
\\
s &:=  \tuple{\tuple{\textsf{p}_\textsf{s}, t_m}, \tuple{\textsf{s},k_m,t_m+1}, \tuple{\textsf{s},k_m,t_m+2}, \ldots}.
\end{align*}
We now show that $\trace \circ f = \xi$ and $f(\pi_\sw) = \pi_\A$, from which the theorem follows directly.
Fix $p = \tuple{t^m, k^m} \in \Theta_\sw$. 
Since the trace of a concatenation equals the concatenation of the traces,
\begin{align*}
\trace \circ f(p) &= \trace(u_1) \conc \trace(u_2) \conc \ldots \conc \trace(u_{m-1}) \conc \trace(s) 
\\
&= k_1^{[t_2-t_1]} \conc k_2^{[t_3-t_2]} \conc \ldots \conc k_2^{[t_m-t_{m-1}]} \conc k_m^{[\omega]} = \xi(p).
\end{align*}
which establishes the first part. Second, we need to show that $\pi_\A$ and $f(\pi_\sw)$ assign the same probability to all events. Since $\pi_\sw$ has countable support, so has $f(\pi_\sw)$. By construction $f$ is injective, so the preimage of $f(p)$ equals $\set{p}$, and hence $f(\pi_\sw)(\set{f(p)}) = \pi_\sw(p)$. Therefore it suffices to show that $\pi_\A(\set{f(p)}) = \pi_\sw(p)$ for all $p \in \Theta_\sw$. Let $\wstate^\omega = f(p)$, and define $u_i$ and $s$ for this $p$ as above. Then
\begin{align*}
\pi_\A(\wstate^\omega) 
&= \pi_\A(\tuple{\textsf{p}, 0}) \del{\prod_{i=1}^{m-1} \pi_\A(u_i|u^{i-1})} \pi_\A(s|u^{m-1})
\end{align*}
Note that
\begin{align*}
\pi_\A(s|u^{m-1}) &= (1-\theta) \pik(k_i)
\\
\pi_\A(u_i|u^{i-1}) &= 
\theta \pik(k_i) \del{\prod_{j=t_i+1}^{t_{i+1}-1} \pit(\rv Z > j | \rv Z \ge j)} \pit(\rv Z = t_{i+1} | \rv Z \ge t_{i+1}).
\intertext{The product above telescopes, so that}
\pi_\A(u_i|u^{i-1}) &= 
\theta \pik(k_i) \pit(\rv Z = t_{i+1} | \rv Z \ge t_{i+1}).
\end{align*}
We obtain
\begin{align*}
\pi_\A(\wstate^\omega) 
&= 1 \cdot \theta^{m-1} \del{\prod_{i=1}^{m-1} \pik(k_i) \pit(t_{i+1} | t_{i+1} > t_i)} (1-\theta) \pik(k_m)
\\
&= \theta^{m-1}(1-\theta) \pik(k_1) \prod_{i=2}^{m} \pik(k_i) \pit(t_i | t_i > t_{i-1})
\\
&= \pi_\sw(p),
\end{align*}
under the assumption that $\pim$ is geometric with parameter $\theta$.
\end{proof}

\subsubsection{A Loss Bound}\label{sec:switchbound}
We derive a loss bound of the same type as the bound for the
fixed share algorithm (see \secref{sec:fixedshare}).
\begin{theorem}\label{thm:swbound} Fix data $x^n$. Let $\hat\theta = \tuple{t^m, k^m}$ maximise the likelihood $P_{\xi(\hat\theta)}(x^n)$ among all switch parameters of length $m$. Let $\pim(n) = 2^{-n}$, $\pit(n) = 1/(n(n+1))$ and $\pik$ be uniform. Then the loss overhead $-\log P_\sw(x^n) + \log P_{\xi(\hat\theta)}(x^n)$ of the switch distribution is bounded by
\[
m + m\log \card{\Xi} + \log \binom{t_m+1}{m} + \log (m!).
\]
\end{theorem}

\begin{proof}We have
\begin{eqnarray}
&&-\log P_\sw(x^n) +\log P_{\xi(\hat\theta)}(x^n)\nonumber\\
& \le &-\log \pi_\text{sw}(\hat\theta)\nonumber\\
& = &-\log\left(\pim(m)\pik( k_1)\prod_{i=2}^m\pit( t_i|
  t_i> t_{i-1})\pik( k_i)\right)\nonumber\\
& = &-\log\pim(m)+\sum_{i=1}^m
-\log\pik( k_i)+\sum_{i=2}^m -\log\pit( t_i| t_i> t_i-1).\label{eq:switchbound}
\end{eqnarray}
The considered prior $\pit(n)=1/(n(n+1))$ satisfies
\[
\pit(t_i|t_i>t_{i-1})={\pit(t_i)\over\sum_{i=t_{i-1}+1}^\infty\pit(i)}={1/(t_i(t_i+1))\over\sum_{i=t_{i-1}+1}^\infty{1\over i}-{1\over i+1}}={t_{i-1}+1\over t_i(t_i+1)}.
\]
If we substitute this in the last term of~\eqref{eq:switchbound}, the
sum telescopes and we are left with
\begin{equation}\label{eq:swloss1}
\underbrace{-\log(t_1+1)}_{=~0}+\log(t_m+1)+\sum_{i=2}^m\log t_i.
\end{equation}
If we fix $t_m$, this expression is maximised if $t_2,\ldots,t_{m-1}$ take
on the values $t_m-m+2,\ldots,t_m-1$, so that~\eqref{eq:swloss1} becomes
\[
\sum_{i=t_m-m+2}^{t_m+1}\log i
=\log\left({(t_m+1)!\over(t_m-m+1)!}\right)
=\log \binom{t_m+1}{m} + \log (m!)
.\]
The theorem follows if we also instantiate $\pim$ and $\pik$ in~\eqref{eq:switchbound}.
\end{proof}

Note that this loss bound is a function of the index of the last
switch $t_m$ rather than of the sample size $n$; this means that in
the important scenario where the number of switches remains bounded in
$n$, the loss compared to the best partition is $O(1)$.

The bound can be tightened slightly by using the fact that we allow
for switching to the same expert, as also remarked in
Footnote~\ref{fn:reflex} on page~\pageref{fn:reflex}. If we take this
into account, the $m\log\card{\Xi}$ term can be reduced to
$m\log(\card{\Xi}-1)$. If we take this into account, the bound
compares quite favourably with the loss bound for the fixed share
algorithm (see \secref{sec:fixedshare}).  We now investigate how much
worse the above guarantees are compared to those of fixed share. The
overhead of fixed share \eqref{eq:fixedshare.lossbound} is bounded
from above by $n H(\alpha) + m \log(\card{\Xi}-1)$. We first underestimate
this worst-case loss by substituting the optimal value $\alpha = m/n$,
and rewrite
\[
n H(\alpha) 
~\ge~
n H(m/n) 
~\ge~
\log \binom{n}{m}.
\]
Second we overestimate the loss of the switch distribution by substituting the worst case $t_m = n-1$. We then find the maximal difference between the two bounds to be
\begin{multline}
\del{m + m\log(\card{\Xi}-1) + \log \binom{n}{m} + \log (m!)}
-
\del{\log \binom{n}{m}  + m \log(\card{\Xi}-1)}
\\
= m + \log (m!) ~\le~ m + m \log m.
\end{multline}

Thus using the switch distribution instead of fixed share lowers the guarantee by at most $m + m \log m$ bits, which is significant only if the number of switches is relatively large. On the flip side, using the switch distribution does not require any prior knowledge about any parameters. This is a big advantage in a setting where we desire to maintain the bound sequentially. This is impossible with the fixed share algorithm in case the optimal value of $\alpha$ varies with $n$.

\subsubsection{MAP Estimation}\label{sec:switch.map}
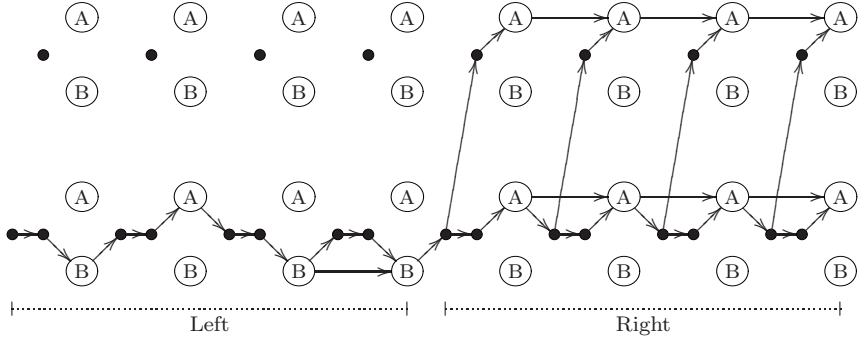
\begin{figure}
\[\xymatrix@R=.5em@C=.5em{
&& \ex{a} &
&& \ex{a} &
&& \ex{a} &
&& \ex{a} &
&& \ex{a} \ar[rrr] &
&& \ex{a} \ar[rrr]&
&& \ex{a} \ar[rrr]&
&& \ex{a}
\\
& \si  &&
& \si  &&
& \si  &&
& \si  &&
& \si \ar[ur] &&
& \si \ar[ur] &&
& \si \ar[ur] &&
& \si \ar[ur] &
\\
&& \ex{b} &
&& \ex{b} &
&& \ex{b} &
&& \ex{b} &
&& \ex{b} &
&& \ex{b} &
&& \ex{b} &
&& \ex{b}
\\
\\
\\
&& \ex{a} &
&& \ex{a} \ar[dr]&
&& \ex{a} &
&& \ex{a} &
&& \ex{a} \ar[dr]\ar[rrr]&
&& \ex{a} \ar[dr]\ar[rrr]&
&& \ex{a} \ar[dr]\ar[rrr]&
&& \ex{a}
\\
\si \ar[r] &  \si \ar[dr] &&
\si \ar[r] &  \si  \ar[ur] &&
\si \ar[r] &  \si  \ar[dr] &&
\si \ar[r] &  \si  \ar[dr] &&
\si \ar[uuuuur]\ar[r] &  \si \ar[ur] &&
\si \ar[uuuuur]\ar[r] &  \si \ar[ur] &&
\si \ar[uuuuur]\ar[r] &  \si \ar[ur] &&
\si \ar[uuuuur]\ar[r] &  \si \ar[ur] &&
\\
&& \ex{b} \ar[ur]&
&& \ex{b} &
&& \ex{b} \ar[ur]\ar[rrr]&
&& \ex{b} \ar[ur]&
&& \ex{b} &
&& \ex{b} &
&& \ex{b} &
&& \ex{b}
\\
\ar@{|.|}_{\text{Left}}[]+0;[rrrrrrrrrrr]+0&&&&&&&&&&&&\ar@{|.|}_{\text{Right}}[]+0;[rrrrrrrrrrr]+0&&&&&&&&&&&
}\]
\caption{MAP estimation for the switch distribution. The sequences of states that can be obtained by following the arrows are exactly those that produce expert sequence \expname{abaabbbb}.}\label{fig:MAP.example}
\end{figure}

The particular nature of the switch distribution allows us to perform MAP estimation efficiently. The MAP sequence of experts is:
\[ \argmax_{\xi^n} P(x^n, \xi^n).\]
We observed in \secref{sec:algo} that Viterbi can be used on unambiguous HMMs. However, the switch HMM is ambiguous, since a single sequence of experts is produced by multiple sequences of states. Still, it turns out that for the switch HMM we can jointly consider all these sequences of states efficiently. Consider for example the expert sequence \expname{abaabbbb}. The sequences of states that produce this expert sequence are exactly the runs through the pruned HMM shown in \figref{fig:MAP.example}. Runs through this HMM can be decomposed in two parts, as indicated in the bottom of the figure. In the right part a single expert is repeated, in our case expert \expname d. The left part is contained in the unstable (lower) band. To compute the MAP sequence we proceed as follows. We iterate over the possible places of the transition from left to right, and then optimise the left and right segments independently. 

In the remainder we first compute the probability of the MAP expert sequence instead of the sequence itself. We then show how to compute the MAP sequence from the fallout of the probability computation. 

To optimise both parts, we define two functions $L$ and $R$. 
\begin{align}
L_i &:= \max_{\xi^i} P(x^i, \xi^i, \tuple{\textsf{p}, i})\\
R_i(\xi) &:= P(x^n, \xi_i = \ldots = \xi_n =\xi | x^{i-1}, \tuple{\textsf{p}, i-1})
\end{align}
Thus $L_i$ is the probability of the MAP expert sequence of length $i$. The requirement $\tuple{\textsf{p}, i}$ forces all sequences of states that realise it to remain in the unstable band. $R_i(\xi)$ is the probability of the tail $x_i, \ldots, x_n$ when expert $\xi$ is used for all outcomes, starting in state $\tuple{\textsf{p}, i-1}$. Combining $L$ and $R$, we have
\[
\max_{\xi^n} P(x^n, \xi^n) = \max_{i \in [n],\xi}  L_{i-1} R_i(\xi).
\]

\paragraph{Recurrence}
$L_i$ and $R_i$ can efficiently be computed using the folowing recurrence relations. First we define auxiliary quantities
\begin{align}
\label{eq:li}
L'_i(\xi) &:= \max_{\xi^i} P(x^i, \xi^i, \tuple{\textsf{u}, \xi, i})
\\
R'_i(\xi) &:= P(x^n, \xi_i = \ldots = \xi_n = \xi | x^{i-1}, \tuple{\textsf{u}, \xi, i})
\end{align}
Observe that the requirement $\tuple{\textsf{u}, \xi, i}$ forces $\xi_i = \xi$. First, $L'_i(\xi)$ is the MAP probability for length $i$ under the constraint that the last expert used is $\xi$. Second, $R'_i(\xi)$ is the MAP probability of the tail $x_i, \ldots, x_n$ under the constraint that the same expert is used all the time. Using these quantities, we have (using the $\gamma_{(\cdot)}$ transition probabilities shown in \eqref{eq:gammas})
\begin{align}
L_i &= \max_\xi L'_i(\xi) \gamma_1
&
R_i(\xi) &= \gamma_2 R'_i(\xi) + \gamma_3 P_\xi(x^n|x^{i-1}).
\end{align}
For $L'_i(\xi)$ and $R'_i(\xi)$ we have the following recurrences:
\begin{align}
L_{i+1}(\xi) &= P_\xi(x_{i+1}|x^i) \max \set{ L'_i(\xi) (\gamma_4 + \gamma_1\gamma_5), L_i \gamma_5}
\\
R'_i(\xi) &= P_\xi(x_i|x^{i-1}) \del{ \gamma_1 R_{i+1}(\xi) + \gamma_4 R'_{i+1}(\xi)}.
\end{align}
The recurrence for $L$ has border case $L_0 = 1$. The recurrence for $R$ has border case $R_n = 1$. 
\begin{equation}\label{eq:gammas}
\begin{aligned}
\gamma_1 &= \wtf{\del{\tuple{\textsf{u}, \xi, i} \to \tuple{\textsf{p}, i}}}
\\
\gamma_2 &=  \wtf{\del{\tuple{\textsf{p}, i-1} \to \tuple{\textsf{p}_\textsf{u}, i-1} \to \tuple{\textsf{u}, \xi, i}}}
\\
\gamma_3 &= \wtf{\del{\tuple{\textsf{p}, i-1} \to \tuple{\textsf{p}_\textsf{s}, i-1} \to \tuple{\textsf{s}, \xi, i}}}
\\
\gamma_4 &= \wtf{\del{\tuple{\textsf{u}, \xi, i} \to \tuple{\textsf{u}, \xi, i+1}}}
\\
\gamma_5 &= \wtf{\del{\tuple{\textsf{p}, i} \to \tuple{\textsf{p}_\textsf{u}, i}\to \tuple{\textsf{u}, \xi, i+1}}}
\end{aligned}
\end{equation}

\paragraph{Complexity}
A single recurrence step of $L_i$ costs $O(\card{\Xi})$ due to the maximisation. All other recurrence steps take $O(1)$. Hence both $L_i$ and $L'_i(\xi)$ can be computed recursively for all $i=1, \ldots, n$ and $\xi \in \Xi$ in time $O(n\card{\Xi})$, while each of $R_i, R'_i(\xi)$ and $P_\xi(x^n|x^{i-1})$ can be computed recursively for all $i=n, \ldots, 1$ and $\xi \in \Xi$ in time $O(n \card{\Xi})$ as well. Thus the MAP probability can be computed in time $O(n \card{\Xi})$. Storing all intermediate values costs $O(n \card{\Xi})$ space as well. 
\paragraph{The MAP Expert Sequence}
As usual in Dynamic Programming, we can retrieve the final solution --- the MAP expert sequence --- from these intermediate values. We redo the computation, and each time that a maximum is computed we record the expert that achieves it. The experts thus computed form the MAP sequence.

\subsection{Run-length Model}
Run-length codes have been used extensively in the context of data
compression, see e.g.~\cite{Moffat2002}. Rather than applying run
length codes directly to the observations, we reinterpret the
corresponding probability distributions as ES-priors, because they may
constitute good models for the distances between consecutive
switches.

The run length model is especially useful if the switches are
clustered, in the sense that some blocks in the expert sequence
contain relatively few switches, while other blocks contain many. The
fixed share algorithm remains oblivious to such properties, as its
predictions of the expert sequence are based on a Bernoulli model: the
probability of switching remains the same, regardless of the index of
the previous switch. Essentially the same limitation also applies to
the universal share algorithm, whose switching probability normally
converges as the sample size increases. The switch distribution
is efficient when the switches are clustered toward the beginning of the sample: its switching probability
decreases in the sample size. However, this may be unrealistic and may
introduce a new unnecessary loss overhead. 

The run-length model is based on the assumption that the
\emph{intervals} between successive switches are independently
distributed according to some distribution $\pit$. After the universal
share model and the switch distribution, this is a third
generalisation of the fixed share algorithm, which is recovered by
taking a geometric distribution for $\pit$. As may be deduced from the
defining HMM, which is given below, we require quadratic running time $O(n^2\card{\Xi})$
to evaluate the run-length model in general.

\subsubsection{Run-length HMM}
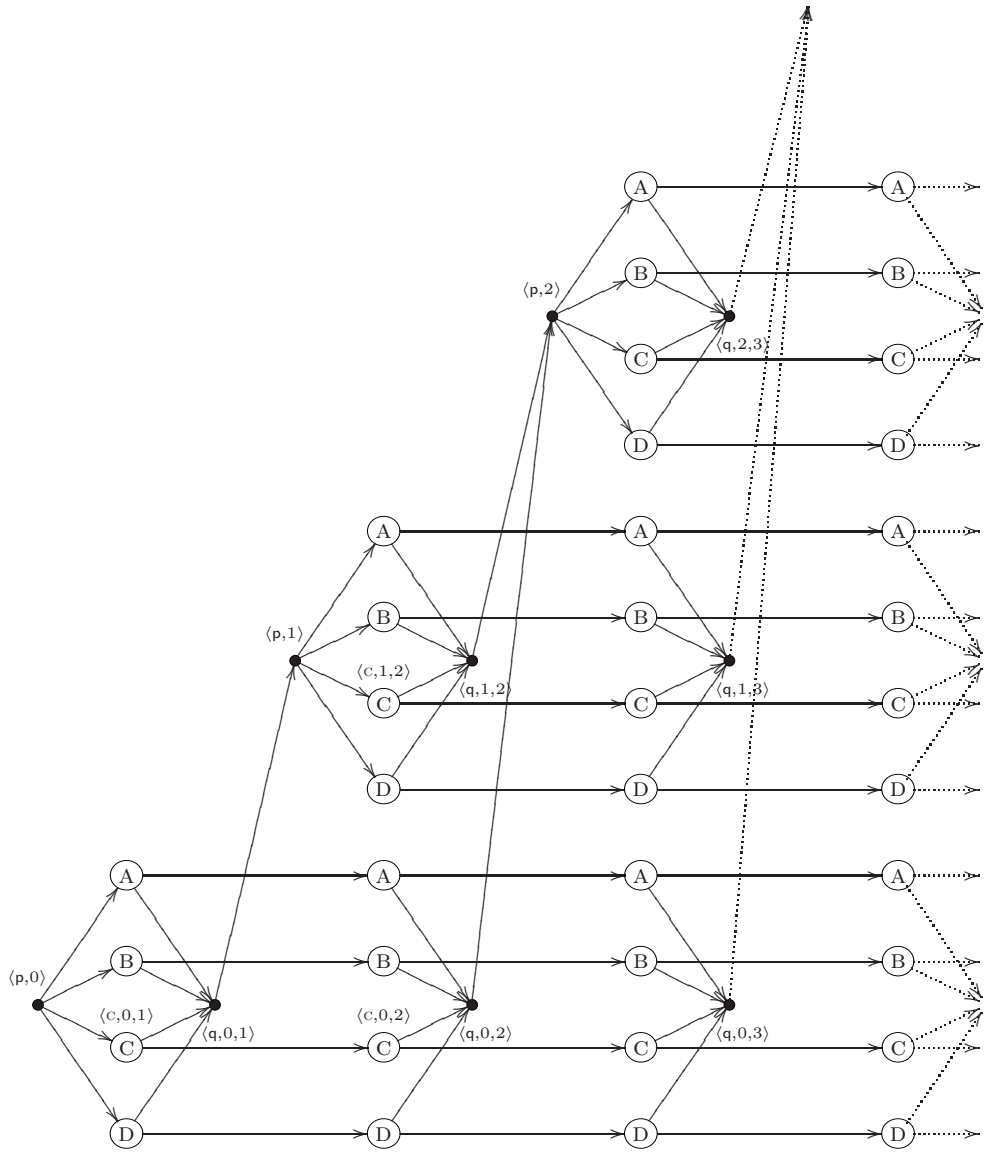
\begin{figure}
\centering
$\xymatrix@R=0.7em{
 &&&
 &&&
 &&&
&&
\\
\\
\\
\\
\\
& &&
& &&
& \ex{a} \ar[rrr] \ar[dddr] &&
& \ex{a} \ar@{.>}[r] \ar@{.>}[dddr] &
\\
\\
& &&
& &&
& \ex{b} \ar[rrr] \ar[dr] &&
& \ex{b} \ar@{.>}[r] \ar@{.>}[dr] &
\\
 &&&
 &&&
\si \name[+<-4pt,10pt>]{\textsf{p},2} \ar[uuur] \ar[ur] \ar[dr] \ar[dddr] &&\si \name[+<5pt,-11pt>]{\textsf{q},2,3} \ar@{.>}[uuuuuuuur]&
 &&
\\
& &&
& &&
& \ex{c} \ar[rrr] \ar[ur] &&
& \ex{c} \ar@{.>}[r] \ar@{.>}[ur] &
\\
\\
& &&
& &&
& \ex{d} \ar[rrr] \ar[uuur] &&
& \ex{d} \ar@{.>}[r] \ar@{.>}[uuur] &
\\
\\
& &&
& \ex{a} \ar[rrr] \ar[dddr] &&
& \ex{a} \ar[rrr] \ar[dddr] &&
& \ex{a} \ar@{.>}[r] \ar@{.>}[dddr] &
\\
\\
& &&
& \ex{b} \ar[rrr] \ar[dr] &&
& \ex{b} \ar[rrr] \ar[dr] &&
& \ex{b} \ar@{.>}[r] \ar@{.>}[dr] &
\\
 &&&
\si \name[+<-4pt,10pt>]{\textsf{p},1} \ar[uuur] \ar[ur] \ar[dr] \ar[dddr] &&\si \name[+<5pt,-11pt>]{\textsf{q},1,2} \ar[uuuuuuuur]&
 &&\si \name[+<5pt,-11pt>]{\textsf{q},1,3} \ar@{.>}[uuuuuuuuuuuuuuuur]&
 &&
\\
& &&
& \ex{c} \name[+<0pt,12pt>]{\expname c,1,2} \ar[rrr] \ar[ur] &&
& \ex{c} \ar[rrr] \ar[ur] &&
& \ex{c} \ar@{.>}[r] \ar@{.>}[ur] &
\\
\\
& &&
& \ex{d} \ar[rrr] \ar[uuur] &&
& \ex{d} \ar[rrr] \ar[uuur] &&
& \ex{d} \ar@{.>}[r] \ar@{.>}[uuur] &
\\
\\
& \ex{a} \ar[rrr] \ar[dddr] &&
& \ex{a} \ar[rrr] \ar[dddr] &&
& \ex{a} \ar[rrr] \ar[dddr] &&
& \ex{a} \ar@{.>}[r] \ar@{.>}[dddr] &
\\
\\
& \ex{b} \ar[rrr] \ar[dr] &&
& \ex{b} \ar[rrr] \ar[dr] &&
& \ex{b} \ar[rrr] \ar[dr] &&
& \ex{b} \ar@{.>}[r] \ar@{.>}[dr] &
\\
\si \name[+<-4pt,10pt>]{\textsf{p},0} \ar[uuur] \ar[ur] \ar[dr] \ar[dddr] &&\si \name[+<5pt,-11pt>]{\textsf{q},0,1}  \ar[uuuuuuuur]&
 &&\si\name[+<5pt,-11pt>]{\textsf{q},0,2}\ar[uuuuuuuuuuuuuuuur]&
 &&\si\name[+<5pt,-11pt>]{\textsf{q},0,3}\ar@{.>}[uuuuuuuuuuuuuuuuuuuuuuuur]&
 &&
\\
& \ex{c} \name[+<0pt,12pt>]{\expname c,0,1} \ar[rrr] \ar[ur] &&
& \ex{c} \name[+<0pt,12pt>]{\expname c,0,2} \ar[rrr] \ar[ur] &&
& \ex{c} \ar[rrr] \ar[ur] &&
& \ex{c} \ar@{.>}[r] \ar@{.>}[ur] &
\\
\\
& \ex{d} \ar[rrr] \ar[uuur] &&
& \ex{d} \ar[rrr] \ar[uuur] &&
& \ex{d} \ar[rrr] \ar[uuur] &&
& \ex{d} \ar@{.>}[r] \ar@{.>}[uuur] &
}$
\caption{HMM for the run-length model}\label{fig:dfe.automaton}
\end{figure}
Let $\mathbb S := \set{\tuple{m,n} \in \nats^2 \mid m < n}$, and let $\pit$ be a distribution on $\posints$. The specification of the run-length HMM is given using $\wstates = \wstates_\sil \cup \wstates_\prd$ by:
\begin{subequations}
\begin{gather}
\begin{aligned}
\wstates_\sil &= \set{\textsf{q}} \times \mathbb S ~\cup\, \set{\textsf{p}} \times \nats \qquad
&
\wnl(\xi,m,n) &= \xi
\\
\wstates_\prd &= \Xi \times \mathbb S
&
\winit(\textsf{p},0) &= 1
\end{aligned}
\\
\label{eq:dfe.wtf.def}
\wtf \del{
\begin{aligned}
\tuple{\textsf{p},n} & \to \tuple{\xi, n, n+1}
\\
\tuple{\xi, m, n} & \to \tuple{\xi, m, n+1}
\\
\tuple{\xi, m, n} & \to \tuple{\textsf{q}, m, n}
\\
\tuple{\textsf{q}, m, n} & \to \tuple{\textsf{p}, n}
\end{aligned}} =
\del{
\begin{gathered}
w(\xi) 
\\
\pit(\rv Z > n | \rv Z \ge n)
\\
\pit(\rv Z = n | \rv Z \ge n)
\\
1
\end{gathered}}
\end{gather}
\end{subequations}

\subsubsection{A Loss Bound}\label{sec:rlebound}

\begin{theorem}\label{thm:rlebound} Fix data $x^n$. Let $\xi^n$ maximise the likelihood $P_{\xi^n}(x^n)$ among all expert sequences with $m$ blocks. For $i=1, \ldots, m$, let $\delta_i$ and $k_i$ denote the length and expert of block $i$. Let $\pik$ be the uniform distribution on experts, and let $\pit$ be a distribution satisfying $-\log \pit(n) \le \log n + 2 \log\log (n+1) + 3$ (for instance an Elias code). Then the loss overhead $- \log P(x^n) + \log P_{\xi^n}(x^n)$ is bounded by
\[
m\del{\log|\Xi|
+
\log {n\over m}
+
2 \log \log \del{\frac{n}{m}+1} 
+ 3}.
\]
\end{theorem}
\begin{proof} We overestimate
\begin{eqnarray}
&&-\log P_\text{rl}(x^n)-(-\log P_{\xi^n}(x^n)) \nonumber\\
& \le &-\log \pi_\text{rl}(\xi^n) \nonumber\\
& = &\sum_{i=1}^m-\log\pik(k_i)
     +\sum_{i=1}^{m-1}-\log \pit(\rv Z=\delta_i)-\log\pit(\rv Z \ge \delta_m) \nonumber\\
\label{eq:rle.bound.1}
& \le & \sum_{i=1}^m-\log\pik(k_i) + \sum_{i=1}^m-\log\pit(\delta_i).
\end{eqnarray}
Since $-\log \pit$ is concave, by Jensen's inequality we have
\[
\sum_{i=1}^m{1\over m}\cdot-\log\pit(\delta_i)\le -\log\pit\left({1\over m}\sum_{i=1}^m
  \delta_i\right)=-\log\pit\left(n\over m\right).
\]
In other words, the block lengths are all equal in the worst case. Plugging this into \eqref{eq:rle.bound.1} we obtain
\[
\sum_{i=1}^m-\log\pik(k_i) + m\cdot-\log\pit\left({n\over m}\right).
\]
The result follows by expanding $\pit$ and $\pik$.
\end{proof}

We have introduced two new models for switching: the switch distribution and the run-length model. It is natural to wonder which model to apply. One possibility is to compare asymptotic loss bounds. To compare the bounds given by Theorems~\ref{thm:swbound} and \ref{thm:rlebound}, we substitute $t_m+1=n$ in the bound for the switch distribution. The next step is to determine which bound is better depending on how fast $m$ grows as a function of $n$. It only makes sense to consider $m$ non-decreasing in $n$.

\begin{theorem}
The loss bound of the switch distribution (with $t_n+1=n$) is asymptotically lower than that of the run-length model if $m=o\big(\del{\log n}^2\big)$, and asymptotically higher if $m = \Omega\big(\del{\log n}^2\big)$.\footnote{Let $f,g: \nats \to \nats$. We say $f = o(g)$ if $\lim_{n \to \infty} f(n)/g(n) = 0$. We say $f = \Omega(g)$ if $\exists c > 0 \exists n_0 \forall n \ge n_0: f(n) \ge c g(n)$.}
\end{theorem}
\begin{proof}[Proof sketch]
After eliminating common terms from both loss bounds, it remains to compare
\[
m + m \log m \quad\text{to}\quad  2 m \log \log \del{\frac{n}{m}+1} + 3.
\]
If $m$ is bounded, the left hand side is clearly lower for sufficiently large $n$. Otherwise we may divide by $m$, exponentiate, simplify, and compare
\[
 m  \quad\text{to}\quad \del{\log n - \log m}^2,
\]
from which the theorem follows directly.
\end{proof}
For finite samples, the switch distribution can be used in case the switches are expected to occur early on average, or if the running time is paramount. Otherwise the run-length model is preferable.

\subsubsection{Finite Support}
We have seen that the run-length model reduces to fixed share if the prior on switch distances $\pit$ is geometric, so that it can be evaluated in linear time in that case. We also obtain a linear time algorithm when $\pit$ has finite support, because then only a constant number of states can receive positive weight at any sample size. For this reason it can be advantageous to choose a $\pit$ with finite support, even if one expects that arbitrarily long distances between consecutive switches may occur. Expert sequences with such longer distances between switches can still be represented with a truncated $\pit$ using a sequence of switches from and to the same expert. This way, long runs of the same expert receive exponentially small, but positive, probability.

\section{Extensions}\label{sec:loose.ends}
The approach described in Sections~\ref{sec:es.priors} and
\ref{sec:hmms} allows efficient evaluation of expert models that can
be defined using small HMMs. It is natural to look for additional
efficient models for combining experts that cannot be expressed as
small HMMs in this way.

In this section we describe a number of such extensions to the model
as described above. In \secref{sec:alg.tricks} we outline different
methods for approximate, but faster, evaluation of large HMMs. The
idea behind \secref{sec:rec.comb} is to treat a \emph{combination} of
experts as a single expert, and subject it to ``meta'' expert
combination. Then in \secref{sec:conditioning.on.data} we outline a
possible generalisation of the considered class of HMMs, allowing the
ES-prior to depend on observed data. Finally we propose an
alternative to MAP expert sequence estimation that is efficiently
computable for general HMMs.

\subsection{Fast Approximations}\label{sec:alg.tricks}
For some applications, suitable ES-priors do not admit a description
in the form of a small HMM. Under such circumstances we might require
an exponential amount of time to compute quantities such as the
predictive distribution on the next expert \eqref{eq:predictive}. For
example, although the size of the HMM required to describe the
elementwise mixtures of \secref{sec:mixtures} grows only polynomially
in $n$, this is still not feasible in practice. Consider that the
transition probabilities at sample size $n$ must depend on the number
of times that each expert has occurred previously. The number of
states required to represent this information must therefore be at
least ${n+k-1\choose k-1}$, where $k$ is the number of experts. For
five experts and $n=100$, we already require more than four million
states! In the special case of mixtures, various methods exist to
efficiently find good parameter values, such as expectation
maximisation, see e.g.\ \cite{McLachlanPeel2000} and Li and Barron's
approach \cite{libarron99}. Here we describe a few general methods to
speed up expert sequence calculations.

\subsubsection{Discretisation}
The simplest way to reduce the running time of \algref{alg:forward} is to reduce the number of states of the input HMM, either by simply omitting states or by identifying states with similar futures. This is especially useful for HMMs where the number of states grows in $n$, e.g.\ the HMMs where the parameter of a Bernoulli source is learned: the HMM for universal elementwise mixtures of \figref{graph:unimix} and the HMM for universal share of \figref{graph:universal.share}. At each sample size $n$, these HMMs contain states for count vectors $(0,n), (1,n-1), \ldots, (n,0)$. In \cite{Jaakkola2003} Monteleoni and Jaakkola manage to reduce the number of states to $\sqrt n$ when the sample size $n$ is known in advance. We conjecture that it is possible to achieve the same loss bound by joining ranges of well-chosen states into roughly $\sqrt n$ super-states, and adapting the transition probabilities accordingly.

\subsubsection{Trimming}
Another straightforward way to reduce the running time of \algref{alg:forward} is by run-time modification of the HMM. We call
this \emph{trimming}. The idea is to drop low probability transitions
from one sample size to the next. For example, consider the HMM
for elementwise mixtures of two experts, \figref{graph:unimix}. The
number of transitions grows linearly in $n$, but depending on the
details of the application, the probability mass may concentrate on a
subset that represents mixture coefficients close to the optimal
value. A speedup can then be achieved by always retaining only the
smallest set of transitions that are reached with probability $p$, for
some value of $p$ which is reasonably close to one. The lost
probability mass can be recovered by renormalisation. 

\subsubsection{The ML Conditioning Trick}
A more drastic approach to reducing the running time can be applied
whenever the ES-prior assigns positive probability to all expert
sequences. Consider the desired marginal probability~\eqref{eq:marg}
which is equal to:
\begin{equation}\label{eq:condtrickmarg}
  P(x^n)=\sum_{\xi^n\in\Xi^n}\pi(\xi^n)\joint(x^n~|~\xi^n).
\end{equation}
In this expression, the sequence of experts $\xi^n$ can be interpreted
as a parameter. While we would ideally compute the Bayes marginal
distribution, which means integrating out the parameter under the
ES-prior, it may be easier to compute a point estimator for $\xi^n$
instead. Such an estimator $\xi(x^n)$ can then be used to find a lower
bound on the marginal probability:
\begin{equation}\label{eq:bound}
\pi(\xi(x^n))\joint(x^n~|~\xi(x^n))\quad\le\quad\joint(x^n).
\end{equation}
The first estimator that suggests itself is the Bayesian maximum
a-posteriori:
\[\xi_\text{map}(x^n):=
\argmax_{\xi^n\in\Xi^n}\pi(\xi^n)P(x^n~|~\xi^n).\]
In \secref{sec:algo} we explain that this estimator is generally hard to
compute for ambiguous HMMs, and for unambiguous HMMs it is as hard as
evaluating the marginal \eqref{eq:condtrickmarg}. One estimator that is
much easier to compute is the maximum likelihood (ML) estimator, which
disregards the ES-prior $\pi$ altogether:
\[\xi_\text{ml}(x^n):=\argmax_{\xi^n\in\Xi^n}P(x^n~|~\xi^n).\]
The ML estimator may correspond to a much smaller term
in~\eqref{eq:condtrickmarg} than the MAP estimator, but it has the
advantage that it is extremely easy to compute. In fact, letting
$\hat\xi^n:=\xi_\text{ml}(x^n)$, each expert $\hat\xi_i$ is a function
of only the corresponding outcome $x_i$. Thus, calculation of the ML
estimator is cheap. Furthermore, if the goal is not to find a lower
bound, but to predict the outcomes $x^n$ with as much confidence as
possible, we can make an even better use of the estimator if we use it
sequentially. Provided that $\joint(x^n)>0$, we can approximate:
\begin{equation}\label{eq:approx}
  \begin{split}
    P(x^n) = \prod_{i=1}^nP(x_i| x^{i-1}) 
    &= \prod_{i=1}^n\sum_{\xi_i\in\Xi}\joint(\xi_i| x^{i-1})P_{\xi_i}(x_i| x^{i-1})
    \\
    & \approx \prod_{i=1}^n\sum_{\xi_i\in\Xi}\pi(\xi_i|\hat\xi^{\,i-1})P_{\xi_i}(x_i|
    x^{i-1}) =: \tilde P(x^n).
  \end{split}
\end{equation}
This approximation improves the running time if the conditional
distribution $\pi(\xi_n|\xi^{n-1})$ can be computed more efficiently
than $P(\xi_n | x^{n-1})$, as is often the case.

\begin{example} As can be seen in \figref{fig:zoology}, the running
  time of the universal elementwise mixture model (cf.\ \secref{sec:mixtures}) is
  $O(n^{\card{\Xi}})$, which is prohibitive in practice, even for
  small $\Xi$. We apply the above approximation. For simplicity we impose the
  uniform prior density $w(\alpha) = 1$ on the mixture coefficients. 
We use the generalisation of Laplace's Rule of Succession
to multiple experts, which states:
\begin{equation}
\pi_\text{ue}(\xi_{n+1}|\xi^n) 
~=~ \int_{\simplex(\Xi)} \alpha(\xi_{n+1}) w(\alpha | \xi^n)  \dif
\alpha
~=~ \frac{ \card{\set{j \le n \mid \xi_j = \xi_{n+1}}}+1 }{n + \card{\Xi}}.
\end{equation}
Substitution in \eqref{eq:approx} yields the following predictive distribution:
\begin{equation}
\begin{split}
\tilde P(x_{n+1} | x^n) 
&= \sum_{\xi_{n+1}\in\Xi}\pi(\xi_{n+1}~|~\hat\xi^{\,n})P_{\xi_{n+1}}(x_{n+1}~|~
    x^{n})
\\
&= \sum_{\xi_{n+1}} \frac{ |\{j \le n
      \mid \hat\xi_j(x^n) = \xi_{n+1}\}|+1 }{n + \card{\Xi}}
P_{\xi_{n+1}}(x_{n+1}|x^n).
\end{split}
\end{equation}
By keeping track of the number of occurrences of each expert in the ML
sequence, this expression can easily be evaluated in time proportional
to the number of experts, so that $\tilde P(x^n)$ can be computed in
the ideal time $O(n\card{\Xi})$. (one has to consider all experts at all sample sizes)
\qedex
\end{example}
The difference between $\joint(x^n)$ and $\tilde\joint(x^n)$ is
difficult to analyse in general, but the approximation does have two
encouraging properties. First, the lower bound~\eqref{eq:bound} on the
marginal probability, instantiated for the ML estimator, also provides
a lower bound on $\tilde\joint$. We have
\[
\tilde P(x^n)\quad\ge\quad\prod_{i=1}^n\pi(\hat\xi_i~|~\hat\xi^{\,i-1})P_{\hat\xi_i}(x_i~|~
   x^{i-1})~=~\pi(\hat\xi^n)\joint(x^n~|~\hat\xi^{\,n}).
\]
To see why the approximation gives higher probability than the bound,
consider that the bound corresponds to a defective distribution,
unlike $\tilde\joint$.

Second, the following information processing argument shows that even
in circumstances where the approximation of the posterior
$\tilde\joint(\xi_i~|~ x^{i-1})$ is poor, the approximation of the predictive
distribution $\tilde\joint(x_i~|~ x^{i-1})$ might be acceptable.
\begin{lemma}\label{lemma:div}
Let $P$ and $Q$ be two mass functions on $\Xi\times{\cal X}$ such that
$P(x|\xi)=Q(x|\xi)$ for all outcomes $\tuple{\xi,x}$. Let $P_\Xi$,
$P_{\cal X}$, $Q_\Xi$ and $Q_{\cal X}$ denote the marginal
distributions of $P$ and $Q$. Then $ D(P_{\cal X}\|Q_{\cal X}) \le D(P_\Xi\|Q_\Xi)$.
\end{lemma}
\begin{proof} 
The claim follows from taking~\eqref{eq:transferbound} in expectation under $P_{\xspace}$:
\[ 
E_{P_{\xspace}} \sbr{ -\log \frac{Q(x)}{P(x)}}
~\le~
E_{P_{\xspace}} E_P \sbr{ -\log \frac{Q(\xi)}{P(\xi)} \bigg| x}
~=~
E_{P_\Xi} \sbr { -\log \frac{Q(\xi)}{P(\xi)}}
. \qedhere\]
\end{proof}
After observing a sequence $x^n$, this lemma, supplied with the
distribution on the next expert and outcome
$\joint(\rv{\xi_{n+1}},\rv{x_{n+1}}~|~ x^n)$, and its approximation
$\pi(\rv{\xi_{n+1}}~|~\xi(x^n))P_{\rv{\xi_{n+1}}}(\rv{x_{n+1}}~|~ x^n)$, shows
that the divergence between the predictive distribution on the next
\emph{outcome} and its approximation, is at most equal to the
divergence between the posterior distribution on the next
\emph{expert} and its approximation. In other words, approximation
errors in the posterior tend to cancel each other out during prediction.

\subsection{Data-Dependent Priors}\label{sec:conditioning.on.data}
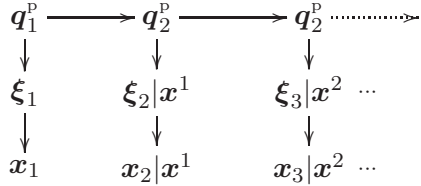
\begin{figure}
\[
\xymatrix@R=1em{
{ \rv \wstate^\prd_1} \ar[r] \ar[d] & {\rv \wstate^\prd_2} \ar[r] \ar[d] & {\rv \wstate^\prd_2} \ar@{.>}[r] \ar[d] & 
\\
{\rv \xi_1} \ar[d] & {\rv \xi_2|\rv x^1} \ar[d] & {\rv \xi_3|\rv x^2} \ar[d]  \ar@{}[r]|\cdots &
\\
{\rv x_1} & {\rv x_2|\rv x^1} & {\rv x_3|\rv x^2} \ar@{}[r]|\cdots &
}
\]
\caption{Conditioning ES-prior on past observations for free}\label{fig:conditioning}
\end{figure}

To motivate ES-priors we used the slogan \emph{we do not understand the data}. When we discussed using HMMs as ES-priors we imposed the restriction that for each state the associated $\Xi$-PFS was independent of the previously produced experts. Indeed, conditioning on the \emph{expert history} increases the running time dramatically as all possible histories must be considered. However, conditioning on the \emph{past observations} can be done \emph{at no additional cost}, as the data are \emph{observed}. The resulting HMM is shown in \figref{fig:conditioning}.
We consider this technical possibility a curiosity, as it clearly violates our slogan. Of course it is equally feasible to condition on some function of the data. An interesting case is obtained by conditioning on the vector of losses (cumulative or incremental) incurred by the experts. This way we maintain ignorance about the data, while extending expressive power: the resulting ES-joints are generally not decomposable into an ES-prior and expert PFSs. An example is the Variable Share algorithm introduced in \cite{HerbsterWarmuth1998}. 

\subsection{An Alternative to MAP Data Analysis}\label{sec:mapalt}
Sometimes we have data $x^n$ that we want to analyse. One way to do this is by computing the MAP sequence of experts. Unfortunately, we do not know how to compute the MAP sequence for general HMMs. We propose the following alternative way to gain in sight into the data. The forward and backward algorithm compute $P(x^i, \wstate^\prd_i)$ and $P(x^n | \wstate^\prd_i, x^i)$. Recall that $\wstate^\prd_i$ is the productive state that is used at time $i$. From these we can compute the a-posteriori probability $P(\wstate^\prd_i | x^n)$ of each productive state $\wstate^\prd_i$. That is, the posterior probability taking the entire future into account. This is a standard way to analyse data in the HMM literature. \cite{rabiner1989} To arrive at a conclusion about experts, we simply project the posterior on states down to obtain the posterior probability $P(\xi_i | x^n)$ of each expert $\xi \in \Xi$ at each time $i=1, \ldots, n$. This gives us a sequence of mixture weights over the experts that we can, for example, plot as a $\Xi \times n$ grid of gray shades. On the one hand this gives us mixtures, a richer representation than just single experts. On the other hand we lose temporal correlations, as we treat each time instance separately. 

\section{Conclusion}
In prediction with expert advice, the goal is to formulate prediction strategies that perform as well as the best possible expert (combination). Expert predictions can be combined by taking a weighted mixture at every sample size. The best combination generally evolves over time. In this paper we introduced expert sequence priors (ES-priors), which are probability distributions over infinite sequences of experts, to model the trajectory followed by the best expert combination. Prediction with expert advice then amounts to marginalising the joint distribution constructed from the chosen ES-prior and the experts' predictions.

We employed hidden Markov models (HMMs) to specify ES-priors. HMMs' explicit notion of current state and state-to-state evolution naturally fit the temporal correlations we seek to model. For reasons of efficiency we use HMMs with silent states. The standard algorithms for HMMs (Forward, Backward, Viterbi and Baum-Welch) can be used to answer questions about the ES-prior as well as the induced distribution on data. The running time of the forward algorithm can be read off directly from the graphical representation of the HMM.

Our approach allows unification of many existing expert models, including mixture models and fixed share. We gave their defining HMMs and recovered the best known running times. We also introduced two new parameterless generalisations of fixed share. The first, called the switch distribution, was recently introduced to improve model selection performance. We rendered its parametric definition as a small HMM, which shows how it can be evaluated in linear time. The second, called the run-length model, uses a run-length code in a novel way, namely as an ES-prior. This model has quadratic running time. We compared the loss bounds of the two models asymptotically, and showed that the run-length model is preferred if the number of switches grows like $\del{\log n}^2$ or faster, while the switch distribution is preferred if it grows slower. We provided graphical representations and loss bounds for all considered models.

Finally we described a number of extensions of the ES-prior/HMM
approach, including approximating methods for large HMMs.

\section*{Acknowledgements}
Peter Gr\"unwald's and Tim van Erven's suggestions significantly improved the quality of this paper. Thank you!

\bibliography{../experts}

\begin{thebibliography}{10}

\bibitem{bousquet2003}
O.~Bousquet.
\newblock A note on parameter tuning for on-line shifting algorithms.
\newblock Technical report, Max Planck Institute for Biological Cybernetics,
  2003.

\bibitem{bousquet2002}
O.~Bousquet and M.~K. Warmuth.
\newblock Tracking a small set of experts by mixing past posteriors.
\newblock {\em Journal of Machine Learning Research}, 3:363--396, 2002.

\bibitem{cover1991}
T.~M. Cover and J.~A. Thomas.
\newblock {\em Elements of Information Theory}.
\newblock John Wiley \& Sons, 1991.

\bibitem{dawid1984}
A.~P. Dawid.
\newblock Statistical theory: The prequential approach.
\newblock {\em Journal of the Royal Statistical Society, Series A}, 147, Part
  2:278--292, 1984.

\bibitem{HerbsterWarmuth1995}
M.~Herbster and M.~K. Warmuth.
\newblock Tracking the best expert.
\newblock In {\em Proceedings of the 12th Annual Conference on Learning Theory
  (COLT 1995)}, pages 286--294, 1995.

\bibitem{HerbsterWarmuth1998}
M.~Herbster and M.~K. Warmuth.
\newblock Tracking the best expert.
\newblock {\em Machine Learning}, 32:151--178, 1998.

\bibitem{libarron99}
J.~Q. Li and A.~R. Barron.
\newblock Mixture density estimation.
\newblock In S.~A. Solla, T.~K. Leen, and K.-R. M{\"u}ller, editors, {\em
  NIPS}, pages 279--285. The MIT Press, 1999.

\bibitem{McLachlanPeel2000}
G.~McLachlan and D.~Peel.
\newblock {\em Finite Mixture Models}.
\newblock Wiley Series in Probability and Statistics, 2000.

\bibitem{Moffat2002}
A.~Moffat.
\newblock {\em Compression and Coding Algorithms}.
\newblock Kluwer Academic Publishers, 2002.

\bibitem{Jaakkola2003}
C.~Monteleoni and T.~Jaakkola.
\newblock Online learning of non-stationary sequences.
\newblock {\em Advances in Neural Information Processing Systems}, 16, 2003.

\bibitem{rabiner1989}
L.~R. Rabiner.
\newblock A tutorial on hidden {M}arkov models and selected applications in
  speech recognition.
\newblock In {\em Proceedings of the {IEEE}}, volume 77, issue 2, pages
  257--285, 1989.

\bibitem{threemusketeers07}
T.~van Erven, P.~D. Gr{\"u}nwald, and S.~de~Rooij.
\newblock Catching up faster in {B}ayesian model selection and model averaging.
\newblock In {\em To appear in Advances in Neural Information Processing
  Systems 20 (NIPS 2007)}, 2008.

\bibitem{volfwillems1998}
P.~Volf and F.~Willems.
\newblock Switching between two universal source coding algorithms.
\newblock In {\em Proceedings of the Data Compression Conference, Snowbird,
  Utah}, pages 491--500, 1998.

\bibitem{Vovk1999}
V.~Vovk.
\newblock Derandomizing stochastic prediction strategies.
\newblock {\em Machine Learning}, 35:247--282, 1999.

\bibitem{xiebarron2000}
Q.~Xie and A.~Barron.
\newblock Asymptotic minimax regret for data compression, gambling and
  prediction.
\newblock {\em IEEE Transactions on Information Theory}, 46(2):431--445, 2000.

\end{thebibliography}

\end{document}